%% file: main.tex
\documentclass[10pt,twocolumn,letterpaper]{article}

\usepackage{cvpr}

\input{preamble}
\input{content_packages}

\definecolor{cvprblue}{rgb}{0.21,0.49,0.74}
\usepackage[pagebackref,breaklinks,colorlinks,allcolors=cvprblue]{hyperref}

\def\confName{CVPR}
\def\confYear{2027}

\title{StereoHand: Geometry-Aware Hand Motion Reconstruction \\ from Egocentric Stereo Videos}

\author{
Kerui Ren$^{1,2}$ \quad
Linning Xu$^{3}$ \quad
Tenghui Wang$^{1,2}$ \quad
Weipeng Zhong$^{1,2}$ \quad
Kaiwen Song$^{4}$ \\
Mulin Yu$^{2\dagger}$ \quad
Tao Lu$^{2\dagger}$ \quad
Bo Dai$^{5}$\\
{\small $^{1}$Shanghai Jiao Tong University, $^{2}$Shanghai Artificial Intelligence Laboratory,} \\
{\small $^{3}$The Chinese University of Hong Kong, $^{4}$University of Science and Technology of China,} \\
{\small $^{5}$The University of Hong Kong}%
}

\begin{document}
\maketitle
\input{figures/teaser}
\input{sec/abstract}
\input{sec/introduction}
\input{sec/related_work}
\input{sec/method}
\input{sec/experiment}

\input{sec/limitation}
\input{sec/conclusion}
\input{sec/statement}
{
    \small
    \bibliographystyle{ieeenat_fullname}
    \bibliography{references_stereohand}
}

\clearpage
\appendix
\input{sec/appendix}

\end{document}

%% file: preamble.tex
\usepackage{multirow}



%% file: content_packages.tex
\usepackage[T1]{fontenc}
\usepackage{amsmath,amssymb,booktabs,graphicx,multirow}
\usepackage{cuted}
\usepackage{tikz}
\usetikzlibrary{arrows.meta,positioning,fit}
\newcommand{\method}{StereoHand}
\newcommand{\SE}{\mathrm{SE}(3)}

%% file: figures/teaser.tex
\begin{strip}
\centering
\vspace{-20pt}
\captionsetup{hypcap=false}
\includegraphics[width=\textwidth]{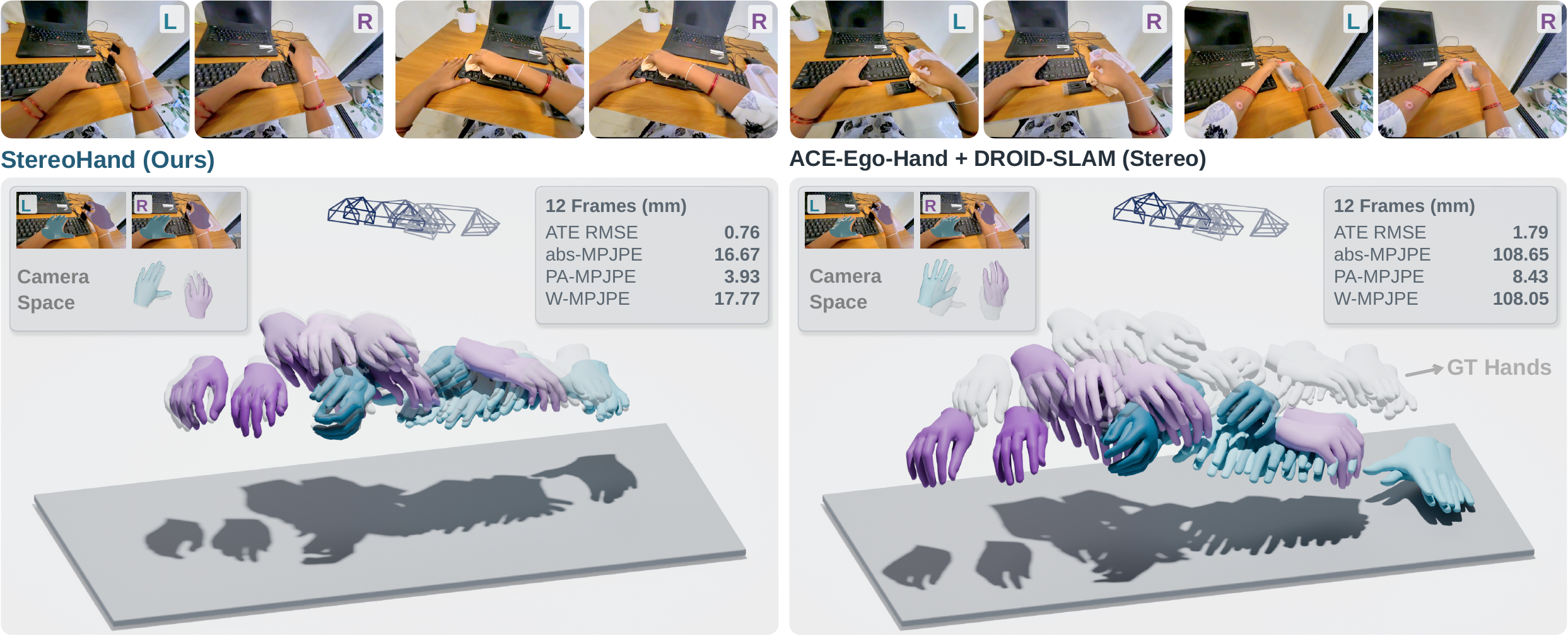}
\captionof{figure}{StereoHand couples stereo-guided metric hand reconstruction with dense-matching SLAM to recover hand motion in a consistent world frame. Our reconstructions are substantially more accurate than those of ACE-Ego-Hand paired with stereo DROID-SLAM. Project page: 
\href{https://stereohand.github.io/}{\textcolor{magenta}{\textbf{https://stereohand.github.io/}}}.}
\label{fig:teaser}
\end{strip}

%% file: sec/abstract.tex
\begin{abstract}
Recovering metric world-space hand motion from egocentric video provides essential geometric demonstrations for embodied learning. Monocular approaches achieve plausible 2D alignment but incur centimeter-scale errors in absolute 3D position, whereas calibrated stereo remains vulnerable to camera tracking drift that corrupts world-space trajectories. To overcome both limitations, we present \method, a unified framework that jointly resolves metric hand localization and stereo camera tracking through a shared geometric foundation model. Specifically, we fuse geometric and stereo foundation features through cross-view alignment and reliability gating to regress MANO parameters in metric camera coordinates. Simultaneously, dense matching descriptors from the same geometric backbone drive a stereo SLAM backend, constraining local rig bundle adjustment and a global keyframe graph with cross-view and temporal correspondences. Across in-domain and out-of-domain benchmarks, \method{} reduces Abs-MPJPE from 161.05\,mm for ACE-Ego-Hand to 16.24\,mm on GenHumanEgo and lowers ATE RMSE by 90.3\% relative to stereo DROID-SLAM on EgoSuite. Ultimately, by coupling feed-forward geometric priors with explicit stereo constraints, \method{} delivers accurate, globally consistent hand trajectories for embodied learning.
\end{abstract}

%% file: sec/introduction.tex
\section{Introduction}
\label{sec:intro}
Egocentric videos capture everyday hand--object interactions at scale~\citep{ego4d,hot3d}, providing a rich source of human demonstrations for embodied learning. Recovering hand motion in a shared world coordinate system turns these visual observations into structured 3D representations that support hand--object interaction reconstruction~\citep{hold,arctic,dexycb}, human-to-robot dexterous motion retargeting~\citep{dexmv,anyteleop}, and the training of vision--language--action models~\citep{egovla} and world-action models~\citep{skelwam}. Crucially, all of these downstream applications rely on physically grounded spatial relationships rather than local hand pose alone. For instance, even when finger poses appear visually accurate, a global displacement of just a few centimeters can alter inferred contact points or miss reach targets entirely. Consequently, making world-space hand reconstruction viable for embodied learning requires preserving the true metric geometry.

Despite this requirement, recovering metric 3D geometry from a single camera remains fundamentally ill-posed because perspective projection couples hand scale with depth~\citep{hz}, forcing recent monocular methods~\citep{hamer,wilor,hawor,dynhamr,infinihand} to rely on learned priors that leave 3D localization ambiguous even under strong 2D alignment. On GenHumanEgo, for example, InfiniHand achieves a 7.90\,mm Procrustes-aligned joint error yet incurs a 45.83\,mm absolute error, demonstrating that millimeter-level local accuracy often coexists with centimeter-level positioning errors. While calibrated stereo resolves this depth ambiguity via synchronized views and a known baseline~\citep{hz}, accurate camera-space depth alone cannot ensure reliable world-space motion because camera pose drift propagates directly into global hand trajectories and distorts close-range hand--object interactions, a regime where existing stereo SLAM systems from ORB-SLAM3~\citep{orbslam3} to DROID-SLAM~\citep{droid} still suffer from substantial egocentric tracking errors. Recovering metric world-space hand motion therefore requires jointly resolving both hand placement relative to the camera and camera motion through the scene.

To resolve both uncertainties jointly, we introduce \method, a unified framework that exploits stereo geometry for metric hand reconstruction and camera pose estimation. For hand reconstruction, multi-view features from a geometric foundation model~\citep{pi3} are fused with correspondence cues from a stereo foundation model~\citep{ffs} through cross-view alignment and reliability gating, feeding separate hand-localization and MANO~\citep{mano} reconstruction heads. In parallel, the same geometric foundation model supplies dense matching descriptors to a stereo SLAM frontend, where temporal and cross-view correspondences constrain local tracking under a calibrated rig and extend to a global core--auxiliary keyframe graph. By coupling feed-forward geometric predictions with local and global bundle adjustment, \method{} transforms metric camera-space hands into a consistent world frame using the refined camera trajectory. Across public in-domain and out-of-domain benchmarks, \method{} outperforms both monocular baselines and the stereo POEM-v2~\citep{poem} method in camera-space hand accuracy while reducing ATE (translation RMSE) on EgoSuite from 11.08\,cm for DROID-SLAM to 1.07\,cm, directly yielding substantially lower world-space hand-motion errors.

We summarize our primary contributions as follows:
\begin{itemize}
    \item We introduce \method, a unified stereo framework that reconstructs metric world-space hand motion from egocentric video by jointly addressing hand reconstruction and camera tracking.
    \item We combine features from geometric and stereo foundation models through cross-view alignment and reliability-gated fusion to reconstruct hand motion in metric camera coordinates.
    \item We develop a stereo SLAM system based on dense matching, integrating stereo constraints into local rig bundle adjustment and global keyframe optimization to support precise tracking of egocentric camera motion.
\end{itemize}

%% file: sec/related_work.tex
\section{Related Work}
\label{sec:related}
\subsection{Hand Motion Reconstruction}
InterHand2.6M~\citep{interhand} addresses interacting-hand pose estimation, while METRO~\citep{metro} and Mesh Graphormer~\citep{meshgraphormer} recover meshes with transformers. HandOccNet~\citep{handoccnet} targets occlusion robustness and MobRecon~\citep{mobrecon} prioritizes efficient monocular reconstruction. HaMeR~\citep{hamer} scales transformer-based mesh regression, while WiLoR~\citep{wilor} integrates hand localization with reconstruction in unconstrained images. For video sequences, HaWoR~\citep{hawor} pairs camera-space hand estimation with egocentric SLAM and motion infilling, and Dyn-HaMR~\citep{dynhamr} optimizes interacting hands using camera estimates and motion priors. ViDiHand~\citep{vidihand} and ACE-Ego-Hand~\citep{aceegohand} exploit video diffusion priors for coherent and occlusion-robust recovery. MINT~\citep{mint} and InfiniHand~\citep{infinihand} jointly estimate hands and camera motion in world space. Calibrated multi-view methods such as POEM-v2~\citep{poem} aggregate observations via point-based representations.

\subsection{Stereo Depth Estimation}
Stereo matching recovers metric depth from disparity and known camera calibration. RAFT-Stereo~\citep{raftstereo} iteratively refines correspondences via multi-level recurrent updates, while IGEV-Stereo~\citep{igevstereo} combines geometry encoding with iterative disparity refinement. To improve zero-shot generalization, FoundationStereo~\citep{foundationstereo} leverages large-scale synthetic training and foundation-model priors. Fast-FoundationStereo~\citep{ffs} accelerates inference through backbone distillation and streamlined refinement, while Lite Any Stereo~\citep{liteanystereo} and Lite Any Stereo V2~\citep{liteanystereov2} also target efficient zero-shot matching.

\subsection{Stereo Visual Odometry and SLAM}
Stereo visual odometry and SLAM combine stereo geometry with temporal correspondences to recover metric camera motion. ORB-SLAM2~\citep{orbslam2} and ORB-SLAM3~\citep{orbslam3} couple sparse feature reprojection with local bundle adjustment, whereas DROID-SLAM~\citep{droid} learns correspondence updates for joint pose and dense-depth optimization. DPVO~\citep{dpvo} uses sparse patches for efficient learned odometry. MAC-VO~\citep{macvo} models metric-aware covariance for keypoint selection and residual weighting. Geometric foundation models offer complementary priors. DUSt3R~\citep{dust3r} predicts pairwise pointmaps, MASt3R~\citep{mast3r} adds matching descriptors, and VGGT~\citep{vggt} and Pi3~\citep{pi3} estimate multi-view geometry. MASt3R-SLAM~\citep{mast3rslam} builds tracking around two-view pointmaps, while M3~\citep{m3} integrates dense matching into a multi-view foundation backbone.

%% file: sec/method.tex
\section{Method}
\label{sec:method}
\begin{figure*}[t]
\centering
\includegraphics[width=\linewidth]{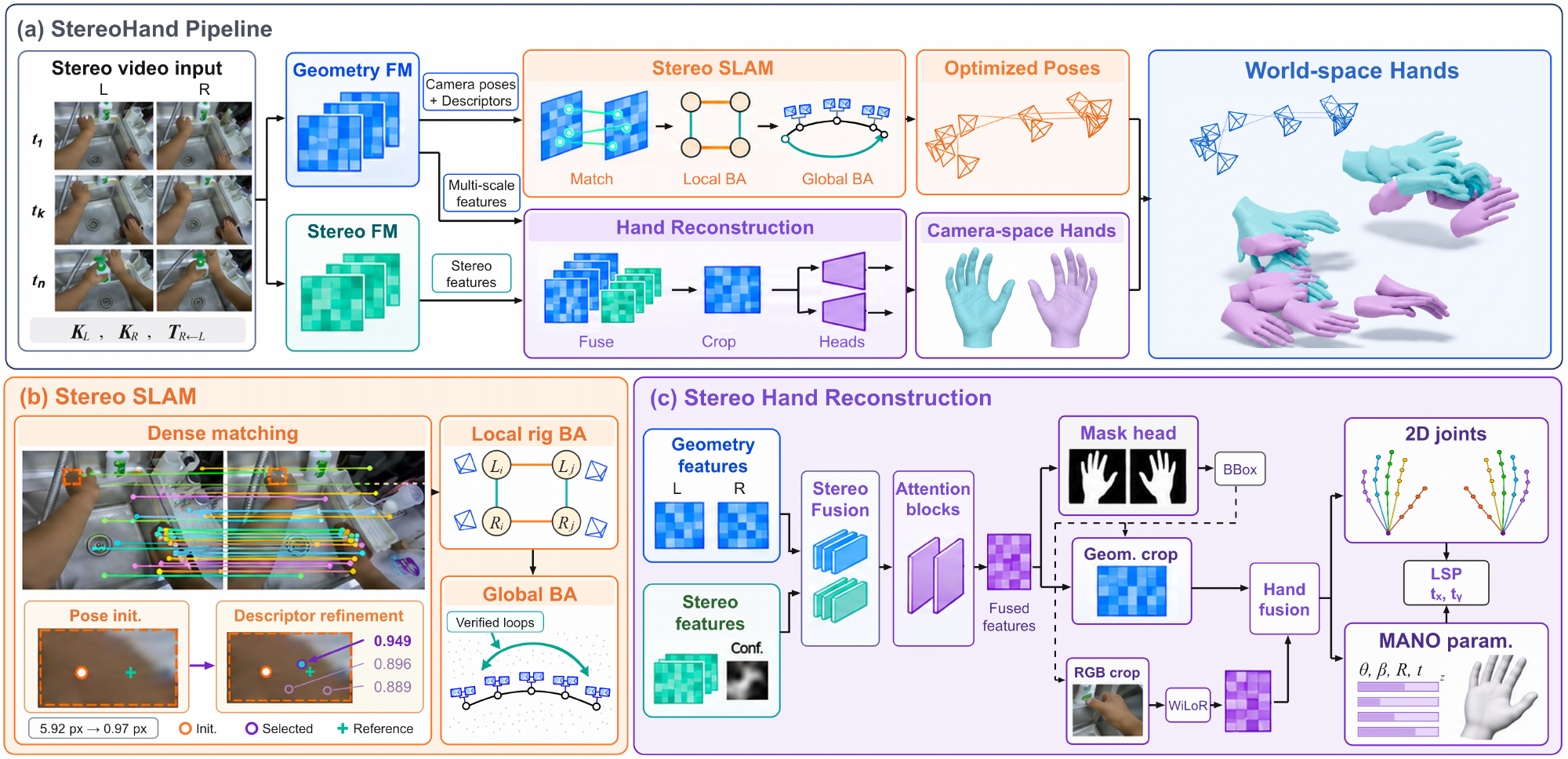}
\caption{\textbf{Overview of \method.} (a) Stereo tracking and metric camera-space hand reconstruction jointly recover world-space motion. (b) Geometry-guided dense matching supplies stereo-temporal constraints for local rig BA and global optimization with loop closures. (c) Stereo-enhanced features guide localization and fuse with cropped appearance to estimate MANO parameters and metric placement.}
\label{fig:pipeline}
\end{figure*}

Fig.~\ref{fig:pipeline} overviews \method, a unified framework for recovering articulated hand motion and camera trajectories from egocentric stereo video. Given a synchronized stereo sequence $\mathcal I=\{\mathbf I_t^L,\mathbf I_t^R\}_{t=1}^{T}$, intrinsics $\mathbf K_t^L,\mathbf K_t^R$, and a fixed metric transform $\mathbf E_R=\mathbf T_{R\leftarrow L}$, we index timestamps by $t\in\{1,\ldots,T\}$, stereo views by $s\in\{L,R\}$, and hand sides by $h\in\{\mathrm{left},\mathrm{right}\}$. Taking the left camera as the rig reference with $\mathbf E_L=\mathbf I_4$, we estimate camera-to-world poses $\mathbf T_t=(\mathbf R_t,\mathbf t_t)\in\SE$ forming $\mathcal T=\{\mathbf T_t\}_{t=1}^{T}$, where superscripts $c$, $L$, and $W$ denote the virtual crop, left-camera, and world frames. Each hand state consists of MANO~\citep{mano} pose $\boldsymbol{\theta}_t^h\in\mathbb R^{15\times3}$, shape $\boldsymbol{\beta}_t^h\in\mathbb R^{10}$, global orientation $\boldsymbol{\Phi}_t^{W,h}\in\mathbb R^3$, and wrist position $\mathbf w_t^{W,h}\in\mathbb R^3$ with axis-angle rotations. Bold symbols denote vectors, matrices, or feature tensors; scalar variables remain unbolded. In each window of $N$ stereo timestamps, Pi3~\citep{pi3} processes all $2N$ images to extract full-image features $\mathbf F$, from which hand-crop features $\mathbf f$ are derived to reconstruct hands in frame $c$, convert them to frame $L$, and transform them to world space $W$ using $\mathcal T$. Sec.~\ref{sec:hand_method} presents stereo-guided hand localization and reconstruction, and Sec.~\ref{sec:slam_method} introduces dense-matching stereo SLAM with local and global bundle adjustment.

\subsection{Stereo-Guided Metric Hand Reconstruction}
\label{sec:hand_method}
\paragraph{Geometry-aware stereo feature fusion.}
To provide metric depth cues for hand localization, we inject cross-view stereo evidence into the single-view geometric features. Fast-FoundationStereo~\citep{ffs} processes rectified stereo pairs to estimate disparity and stereo features $\mathbf F_{t,\mathrm{stereo}}$. We resample $\mathbf F_{t,\mathrm{stereo}}$ onto the original left-image feature grid and map disparity correspondences back to original image coordinates to obtain a left-to-right mapping $\mathcal C_t$. Left--right consistency and valid image boundaries yield a reliability map $\boldsymbol\alpha_t\in[0,1]^{H_f\times W_f}$ to suppress occlusions and mismatches; $H_f$ and $W_f$ denote the feature-grid height and width.

At each selected Pi3 layer with view features $\mathbf F_{t,\mathrm{geo}}^s$ for $s\in\{L,R\}$, we warp the right features into the left grid via $\widetilde{\mathbf F}_{t,\mathrm{geo}}^R(\mathbf u)=\mathbf F_{t,\mathrm{geo}}^R(\mathcal C_t(\mathbf u))$ and project both views into a shared embedding space as $\mathbf H^L=\psi_{\mathrm{in}}(\mathbf F_{t,\mathrm{geo}}^L)$ and $\mathbf H^R=\psi_{\mathrm{in}}(\widetilde{\mathbf F}_{t,\mathrm{geo}}^R)$. Omitting layer and timestamp indices for clarity, gated stereo fusion proceeds as
\begin{equation}
\begin{aligned}
 \mathbf H_{\mathrm{pair}} &= \mathbf H^L + \boldsymbol\alpha\odot\phi\!\left([\mathbf H^L, \mathbf H^R]\right),\\
 \mathbf H_{\mathrm{attn}} &= \operatorname{CrossAttn}_{\mathcal N}(\mathbf H_{\mathrm{pair}}, \mathbf F_{\mathrm{stereo}}),\\
 \widehat{\mathbf F}_{\mathrm{geo}}^L &= \mathbf F_{\mathrm{geo}}^L + \sigma(g)\,\boldsymbol\alpha\odot\psi_{\mathrm{out}}(\mathbf H_{\mathrm{pair}} + \mathbf H_{\mathrm{attn}}).
\end{aligned}
\label{eq:stereo_fusion}
\end{equation}
Here, $\phi$ fuses channel-concatenated cross-view features, $\operatorname{CrossAttn}_{\mathcal N}$ performs local cross-attention over a stereo neighborhood $\mathcal N$ with invalid neighbors masked out, and $\psi_{\mathrm{out}}$ restores the original channel dimension. A learned sigmoid gate $\sigma(g)$ and the spatial reliability $\boldsymbol\alpha$ jointly modulate the residual update, with $\odot$ denoting element-wise multiplication broadcast across channels, while the mask and MANO branches instantiate this module with separate parameters to support localization and articulated reconstruction independently.

\paragraph{Hand Localization.}
We localize each hand in the left image using the stereo-enhanced geometric features. A Dense Prediction Transformer (DPT) decoder~\citep{dpt} aggregates four intermediate feature maps through channel projection, multi-scale resizing, and coarse-to-fine fusion. A two-channel prediction head upsamples the decoded representation to the input image resolution and produces per-hand probabilities:
\begin{equation}
 \widehat{\mathbf m}_t^h
 =\sigma\!\left(\left[\mathcal H_{\mathrm{mask}}\!\left(
 \operatorname{DPT}(\{\widehat{\mathbf F}_{t,\mathrm{geo},\ell}^{L}\}_{\ell=1}^{4})\right)\right]_h\right),
\label{eq:hand_mask}
\end{equation}
where $h$ selects the hand channel, $\ell$ indexes feature levels, and $\sigma$ is the element-wise sigmoid. Thresholding $\widehat{\mathbf m}_t^h$ yields a tight bounding box, which we expand to a square $\mathbf b_t^h\in\mathbb R^4$ to retain context around the hand. Each box defines a virtual crop camera whose optical axis points along the ray through the box center. Let $\mathbf U_t^h$ rotate the left-camera frame into this crop frame and $\mathbf K_t^{c,h}$ denote its intrinsics. The perspective crop is specified by the homography $\mathbf A_t^h=\mathbf K_t^{c,h}\mathbf U_t^h(\mathbf K_t^L)^{-1}$, providing a common sampling geometry for the RGB image and geometric features.

\paragraph{Metric Hand Reconstruction.}
For each detected hand, we combine stereo-enhanced geometry with local appearance to recover articulated pose and metric 3D placement. Specifically, we resample both the geometric features $\widehat{\mathbf F}_{t,\mathrm{geo}}^L$ and the left image $\mathbf I_t^L$ using the perspective homography $\mathbf A_t^h$, applying a horizontal flip to left hands for handedness canonicalization. A feature adapter aligns the cropped geometric representation with the patch grid of a WiLoR encoder~\citep{wilor} operating on the RGB crop, and a $1\times1$ convolution fuses the outputs into a unified hand feature $\mathbf f_t^h$.

From $\mathbf f_t^h$, a joint head predicts 21 heatmaps to extract initial 2D joints via soft-argmax and 21 joint tokens via heatmap-weighted descriptor pooling, while a learned hand query attends to $\mathbf f_t^h$ to capture global hand context. Bidirectional cross-attention exchanges information between the hand and joint tokens before the joint tokens predict residual 2D joint offsets and the hand token regresses MANO parameters and wrist log-depth. Denoting this decoding module by $\mathcal H_{\mathrm{hand}}$ and omitting indices $t$ and $h$ for clarity, its predictions are
\begin{equation}
 (\widehat{\mathbf p},\boldsymbol\theta,\boldsymbol\beta,
   \boldsymbol\Phi^c,\ell_z)=\mathcal H_{\mathrm{hand}}(\mathbf f),
 \ z=\exp(\ell_z),
\label{eq:hand_prediction}
\end{equation}
where $\widehat{\mathbf p}$ collects 21 normalized crop-space 2D joints $\widehat{\mathbf p}_j\in\mathbb R^2$, and $\boldsymbol\theta$, $\boldsymbol\beta$, and $\boldsymbol\Phi^c$ specify articulation, shape, and crop-frame orientation. After undoing left-hand canonicalization, we map $\widehat{\mathbf p}$ to crop pixels $\widehat{\mathbf p}^{\mathrm{pix}}$ and decode side-specific MANO geometry into oriented, wrist-centered joint vectors $\bar{\mathbf j}_j\in\mathbb R^3$ and vertex vectors $\bar{\mathbf v}_n\in\mathbb R^3$. Keeping the predicted depth $z$ fixed, we recover the lateral wrist coordinates via differentiable least-squares projection (LSP)~\citep{vidihand} as
\begin{equation}
\begin{gathered}
 (a^*,b^*)=\arg\min_{a,b}\sum_{j\in\mathcal V}
 \left\|\pi_{\mathbf K^{c}}\!\left(\bar{\mathbf j}_j+[a,b,z]^\top\right)
       -\widehat{\mathbf p}^{\mathrm{pix}}_j\right\|_2^2, \\
 \mathbf w^c=[a^*,b^*,z]^\top,
\end{gathered}
\label{eq:hand_lsp}
\end{equation}
where $\pi_{\mathbf K^c}$ denotes perspective projection under the crop intrinsics and $\mathcal V$ indexes valid joints with positive crop-camera depth $\bar j_{j,z}+z>0$. Because fixing $z$ makes projection affine in $a$ and $b$, this objective admits a closed-form solution that couples 2D joint evidence with stereo-informed metric depth. Restoring frame and hand indices, we map the mesh, wrist, and orientation into the left-camera frame via
\begin{equation}
\begin{aligned}
 \mathbf v_{t,n}^{L,h}&=(\mathbf U_t^h)^\top(\bar{\mathbf v}_{t,n}^h+\mathbf w_t^{c,h}),\\
 \mathbf w_t^{L,h}&=(\mathbf U_t^h)^\top\mathbf w_t^{c,h},\\
 \operatorname{Rot}(\boldsymbol\Phi_t^{L,h})
   &=(\mathbf U_t^h)^\top\operatorname{Rot}(\boldsymbol\Phi_t^{c,h}),
\end{aligned}
\label{eq:hand_camera}
\end{equation}
where $n$ indexes mesh vertices and $\operatorname{Rot}$ maps axis-angle vectors to rotation matrices. This inverse crop rotation updates global placement while preserving MANO articulation and shape, and the optimized rig trajectory transforms these camera-space hands into world coordinates.

\paragraph{Training and Loss Functions.}
We train the localization and reconstruction branches separately, optimizing their stereo fusion modules and task heads while keeping Pi3 and Fast-FoundationStereo frozen and fine-tuning the WiLoR encoder with the reconstruction branch. Hand localization is supervised over valid annotations by combining binary cross-entropy and Dice losses as
\begin{equation}
 \mathcal L_{\text{mask}}
 =\lambda_{\text{BCE}}\mathcal L_{\text{BCE}}(\widehat{\mathbf m},\mathbf m)
 +\lambda_{\text{Dice}}\mathcal L_{\text{Dice}}(\widehat{\mathbf m},\mathbf m),
\label{eq:mask_loss}
\end{equation}
where $\mathbf m$ denotes the ground-truth binary mask and $\lambda$ coefficients weight the loss terms. Metric hand reconstruction learns from parameter and geometric supervision formulated as:
\begin{equation}
\begin{aligned}
 \mathcal L_{\text{MANO}}
 &=\lambda_{\text{pose}}\mathcal L_{\text{pose}}
   +\lambda_{\text{shape}}\mathcal L_{\text{shape}}
   +\lambda_{\text{orient}}\mathcal L_{\text{orient}} \\
   +\lambda_{\text{trans}}&\mathcal L_{\text{trans}} 
   +\lambda_{\text{reproj}}\mathcal L_{\text{reproj}}
   +\lambda_{\text{3D}}\mathcal L_{\text{3D}}
   +\lambda_{\text{2D}}\mathcal L_{\text{2D}},
\end{aligned}
\label{eq:mano_loss}
\end{equation}
where the first four terms supervise MANO articulation, shape, global orientation, and metric wrist translation, while $\mathcal L_{\text{3D}}$, $\mathcal L_{\text{2D}}$, and $\mathcal L_{\text{reproj}}$ enforce 3D joint, crop-space 2D joint, and full-image reprojection consistency, respectively. Global orientation, wrist translation, and 3D joints are supervised in the left-camera frame. During temporal fine-tuning, we additionally incorporate motion and shape consistency losses alongside detached geometric warps for mask stability, with detailed formulations provided in the supplementary material.

\subsection{Dense Matching for Stereo SLAM}
\label{sec:slam_method}
\paragraph{Geometry-guided Dense Matching.}
We refine the calibrated stereo trajectory using dense correspondences guided by initial metric geometry. Within each window, we backproject point-head depths along calibrated rays using initial camera-head poses, and recover the metric scale from the median ratio of the known stereo baseline to the predicted left--right camera distance. Rescaling translations and pointmaps yields local metric pointmaps $\mathbf X_{t,s}$, while a matching decoder and head concurrently extract unit-normalized descriptors $\mathbf D_t^s$ from the shared features $\mathbf F_{t,\mathrm{geo}}^s$.

Using these metric pointmaps as geometric priors, we project a source pixel $\mathbf u$ from view $s$ at time $i$ into view $s'$ at time $j$ to obtain an initial estimate $\mathbf v_0$, and refine the match within a local neighborhood $\mathcal N_r(\mathbf v_0)$ via
\begin{equation}
 \mathbf v^*=\arg\max_{\mathbf v\in\mathcal N_r(\mathbf v_0)}
 \left\langle \mathbf D_i^s(\mathbf u),\mathbf D_j^{s'}(\mathbf v)\right\rangle.
\label{eq:matching}
\end{equation}
We discard outliers and dynamic regions by enforcing depth and boundary validity, high descriptor similarity, forward--backward cycle consistency, and geometric projection agreement. For temporal matches, the stereo mapping $\mathcal C_t$ additionally verifies four-view rig consistency through $M_{R_i\to R_j}(\mathcal C_i(\mathbf u))\approx\mathcal C_j(M_{L_i\to L_j}(\mathbf u))$, where $M$ denotes a temporal correspondence map coupling the left and right matching paths.

\paragraph{Stereo-temporal Constraints.}
For timestamps $i$ and $j$, temporal factors link $L_i\leftrightarrow L_j$ and $R_i\leftrightarrow R_j$, whereas stereo factors link $L_i\leftrightarrow R_i$ and $L_j\leftrightarrow R_j$. Both views share a common rig pose $\mathbf T_t\in\SE$ via fixed extrinsics $\mathbf E_L=\mathbf I_4$ and $\mathbf E_R=\mathbf T_{R\leftarrow L}$, while a scalar log-scale correction $\eta_{t,s}$ compensates for residual depth scale errors in each pointmap. Given a retained match $m$ between $\mathbf u_m$ and $\mathbf v_m$, the source point transforms into the target view in homogeneous coordinates as
\begin{equation}
 \begin{bmatrix}\mathbf y_m\\1\end{bmatrix}
 =\mathbf E_{s'}\mathbf T_j^{-1}\mathbf T_i\mathbf E_s^{-1}
 \begin{bmatrix}\exp(\eta_{i,s})\mathbf X_{i,s}(\mathbf u_m)\\1\end{bmatrix}.
\label{eq:rig}
\end{equation}
In same-time stereo factors, $\mathbf T_i^{-1}\mathbf T_i$ cancels so the calibrated baseline directly anchors pointmap scale, enabling temporal factors to refine rig motion using scale-consistent geometry while maintaining a fixed physical baseline.

\paragraph{Pose Tracking and Global Optimization.}
Each Pi3 inference jointly processes $N = 1 + N_{\text{his}} + N_{\text{new}}$ stereo pairs ($2N$ images), comprising the current reference keyframe, $N_{\text{his}}$ historical keyframes, and $N_{\text{new}}$ incoming frames. Local BA tracks incoming frames relative to the current reference using stereo and temporal matches. For frames $\mathcal W$ and factors $\mathcal F$, both local and global BA~\citep{ba} minimize
\begin{equation}
 \min_{\{\mathbf T_t,\eta_{t,L},\eta_{t,R}\}_{t\in\mathcal W}}\;
 \sum_{m\in\mathcal F}w_m
       \sum_{k=1}^{3}\rho(r_{m,k}).
\label{eq:ba}
\end{equation}
Here $\mathbf r_m\in\mathbb R^3$ combines two pixel reprojection residuals with a weighted log-depth difference between $\mathbf y_m$ and the scale-corrected target pointmap; $w_m$ is the clipped descriptor cosine similarity and $\rho$ is the Huber penalty. Local BA fixes the initial reference pose while optimizing incoming poses and all per-eye scales; calibration and pointmaps remain fixed.

During factor construction, a frame becomes a keyframe when $\min(c_{\text{src}},c_{\text{tgt}})<\tau_{\text{KF}}$, where $c_{\text{src}}$ and $c_{\text{tgt}}$ measure static matching coverage in the reference and incoming left images. It then becomes the local reference. Selected keyframes form graph nodes connected by temporal and stereo factors.

The last selected keyframe in each window queries SALAD~\citep{salad} for historical keyframes. We jointly rerun Pi3 on the query and retrieved stereo pairs, adding historical matching edges only after geometric verification; verified loop closures supply long-range constraints. Global BA applies Eq.~\ref{eq:ba} to all keyframes and accumulated factors, fixing the first pose to anchor the world frame. Corrections propagate to ordinary frames through their keyframe-relative poses, yielding world-space hands as $\mathbf v_{t,n}^{W,h}=\mathbf R_t\mathbf v_{t,n}^{L,h}+\mathbf t_t$. Further details are provided in the supplementary material.

%% file: sec/experiment.tex
\section{Experiments}
\label{sec:experiments}

\input{tables/baselines}
\input{figures/qualitative}

\input{figures/world_qualitative}

\subsection{Experimental Setup}
\label{sec:experimental_setup}
\paragraph{Datasets \& Metrics.}
We evaluate in-domain performance on 100 EgoSuite~\citep{egosuite} scenes (50 each from EgoPro and EgoStandard) and out-of-domain performance on 50 GenHumanEgo~\citep{genhumanego} scenes. Hand localization is assessed with F1 and frame accuracy (FAcc). Camera-space reconstruction uses Abs-MPJPE, MPJPE, PA-MPJPE, $\mathrm{EPE}_{\text{2D}}$, and Jitter, measuring absolute, wrist-relative, and Procrustes-aligned 3D joint errors, 2D projection error, and temporal smoothness, respectively. Following HaWoR~\citep{hawor}, world-space motion is evaluated using W-MPJPE, WA-MPJPE, RTE, and Accel to assess aligned joint motion, normalized wrist trajectory error, and acceleration error. Camera tracking is measured by SE(3)-aligned ATE without scale fitting and rotational RPE~\citep{sturm}. Detailed evaluation protocols are provided in the supplementary material.

\paragraph{Baselines.}
We benchmark StereoHand against monocular estimators HaMeR~\cite{hamer}, WiLoR~\cite{wilor}, HaWoR~\cite{hawor}, Dyn-HaMR~\cite{dynhamr}, ViDiHand~\cite{vidihand}, MINT~\cite{mint}, ACE-Ego-Hand~\cite{aceegohand}, and InfiniHand~\cite{infinihand}, alongside the stereo-trained POEM-v2~\cite{poem}. Monocular baselines receive the left image and intrinsics, while POEM-v2 and StereoHand utilize both views. In world-space evaluation, MINT and InfiniHand rely on their native monocular trajectories, whereas WiLoR-SLAM, ACE-SLAM, and HaWoR combine saved hand predictions with stereo DROID-SLAM poses. Finally, standalone camera pose comparisons include stereo ORB-SLAM3~\cite{orbslam3}, DROID-SLAM~\cite{droid}, MAC-VO~\cite{macvo}, AirSLAM~\cite{airslam}, and VOCA~\cite{voca}.

\paragraph{Implementation details.}
We train hand localization and reconstruction separately for 300k steps each, using a global batch size of 32 and clips of 8--16 synchronized stereo pairs. Both branches use AdamW~\citep{adamw} with a peak learning rate of $10^{-5}$, weight decay 0.01,  cosine decay to $5\times10^{-7}$, and gradient norm clipping at 1.0. Dense matching is trained separately for 200k steps with a global batch size of 64. Each eye is processed at $378\times518$ pixels. Hand boxes are expanded by $1.5\times$ and resampled to $256\times256$, with the central $256\times192$ region used for hand features. For hand reconstruction, inference uses $(N_{\text{his}},N_{\text{new}})=(2,13)$.

\input{tables/world_hand}
\input{tables/slam_baselines}

\subsection{Camera-Space Results Analysis}
\label{sec:hand_comparison}
As reported in Tab.~\ref{tab:baselines}, StereoHand enhances metric hand placement while preserving accurate articulation and localization, reducing Abs-MPJPE by 22.5\% on EgoSuite and 64.6\% on GenHumanEgo relative to InfiniHand and approximately halving the absolute error of stereo-trained POEM-v2 on GenHumanEgo. On EgoSuite, the gain in absolute error far exceeds the change in PA-MPJPE, underscoring the need to evaluate metric 3D positioning alongside aligned shape. Visually, Fig.~\ref{fig:qualitative} illustrates this contrast across object-holding and reading sequences, where competing methods yield plausible RGB overlays yet their camera X--Z insets expose depth displacement, whereas StereoHand closely matches the ground-truth 3D configurations. Furthermore, whereas diffusion-based baselines such as ACE-Ego-Hand achieve low jitter on GenHumanEgo at the expense of substantial metric drift, StereoHand maintains competitive temporal stability alongside high spatial accuracy by anchoring predictions in stereo geometry.

\subsection{World-Space Results Analysis}
\label{sec:world_comparison}
Improved camera-space hand placement and tracking directly yield more accurate world-space motion, with StereoHand achieving the lowest errors across all four metrics in Tab.~\ref{tab:world_hand} and reducing W-MPJPE by 58.3\% on EgoSuite and 43.9\% on GenHumanEgo relative to InfiniHand. Concurrent gains in WA-MPJPE, RTE, and acceleration error confirm improvements across aligned articulation, wrist trajectories, and temporal dynamics, while outperforming monocular hand estimators paired with stereo DROID-SLAM validates our unified formulation. Visually, Fig.~\ref{fig:world_qualitative} renders ten shared timestamps under identical viewpoints and bounds after a per-method Sim(3) fit to both hands over the common valid clip. Across object-handling and book-manipulation sequences, baselines frequently compress motion paths or distort bimanual spatial layouts, whereas StereoHand faithfully tracks the ground-truth direction and spatial extent, preserving the geometry of two-hand actions.

\subsection{Camera Pose Results Analysis}
\label{sec:pose_estimation}
Accurate world-space hand placement depends on both camera translation and frame-to-frame rotation, and Tab.~\ref{tab:slam_baselines} shows gains in both, with StereoHand achieving the lowest rotation RPE across datasets while cutting ATE relative to stereo DROID-SLAM by 90.3\% on EgoSuite and 12.5\% on GenHumanEgo, where DROID-SLAM already reaches sub-centimeter accuracy. Alongside these SE(3)-aligned quantitative results without scale fitting, Sim(3)-aligned trajectory plots in the supplementary material show that StereoHand faithfully tracks ground-truth turns while baselines often contract or deform. Finally, StereoHand, DROID-SLAM, and MAC-VO cover all 150 sequences, while remaining baselines exhibit tracking failures or large trajectory errors on a subset of scenes.

\subsection{Ablation Study}
\label{sec:ablations}
We conduct ablation studies on 60 scenes randomly sampled from the 150-scene evaluation set, including 30 EgoSuite scenes and 30 GenHumanEgo scenes (Tab.~\ref{tab:ablations}).

For camera-space hands, (1) removing the right view more than doubles Abs-MPJPE, from 17.16 to 40.19\,mm, confirming that cross-view evidence is central to metric placement; (2) removing stereo features degrades both metric and image-space accuracy, showing that stereo cues support depth-aware reconstruction; (3) removing reliability gating has only a minor effect on this subset; and (4) using WiLoR boxes produces the largest reconstruction error, highlighting the importance of accurate localization. 

For world-space hands, (1) removing descriptor refinement degrades both camera tracking and hand trajectories; (2) disabling global BA produces the largest ATE increase, from 0.86 to 2.58\,cm, highlighting its role in suppressing drift; (3) removing loop closure reduces long-range consistency; and (4) removing stereo constraints causes a smaller but consistent degradation, showing that calibrated cross-view geometry complements temporal matching.

\input{tables/ablations}

%% file: tables/baselines.tex
\begin{table*}[t]
\centering
\caption{\textbf{Quantitative comparison of camera-space hand reconstruction} on EgoSuite and GenHumanEgo. StereoHand achieves the lowest absolute joint error on both benchmarks. \textbf{Bold} and \underline{underlined} entries denote the best and second-best results.}
\label{tab:baselines}
\scriptsize
\setlength{\tabcolsep}{2pt}
\renewcommand{\arraystretch}{1.1}
\resizebox{\textwidth}{!}{%
\begin{tabular}{lccccccc@{\hspace{8pt}}ccccccc}
\toprule
& \multicolumn{7}{c}{EgoSuite} & \multicolumn{7}{c}{GenHumanEgo} \\
\cmidrule(lr){2-8}\cmidrule(lr){9-15}
Method & F1 $\uparrow$ & FAcc $\uparrow$ & Abs-MPJPE $\downarrow$ & MPJPE $\downarrow$ & PA $\downarrow$ & $\mathrm{EPE}_{\text{2D}}\downarrow$ & Jitter $\downarrow$ & F1 $\uparrow$ & FAcc $\uparrow$ & Abs-MPJPE $\downarrow$ & MPJPE $\downarrow$ & PA $\downarrow$ & $\mathrm{EPE}_{\text{2D}}\downarrow$ & Jitter $\downarrow$ \\
\midrule
HaMeR & 0.948 & 0.898 & 54.55 & 31.74 & 8.84 & 14.14 & 32.03 & 0.964 & 0.912 & 55.84 & 38.61 & 10.07 & 15.18 & 28.32 \\
WiLoR & 0.987 & 0.962 & 41.34 & 27.36 & 8.12 & 9.75 & 22.76 & 0.985 & 0.961 & 45.31 & 33.14 & 8.76 & 10.65 & 23.19 \\
HaWoR & 0.623 & 0.409 & 167.43 & 67.77 & 24.51 & 69.83 & 49.73 & 0.872 & 0.777 & 109.44 & 39.25 & 14.95 & 29.78 & 43.27 \\
Dyn-HaMR & 0.992 & 0.981 & 514.05 & 21.73 & 7.17 & 8.53 & 48.08 & 0.985 & 0.966 & 119.78 & 22.68 & 10.08 & \underline{8.32} & 26.72 \\
ViDiHand & 0.910 & 0.733 & 163.77 & 44.64 & 16.65 & 33.07 & 8.97 & 0.982 & 0.944 & 214.31 & 33.15 & 10.66 & 17.62 & 8.20 \\
MINT & 0.999 & 0.997 & 74.29 & 25.11 & 8.15 & 11.29 & 19.27 & 0.994 & 0.979 & 50.20 & 21.79 & 8.50 & 11.02 & 17.32 \\
ACE-Ego-Hand & 0.996 & 0.989 & 105.50 & 22.89 & 9.42 & 20.09 & 5.11 & 0.975 & 0.950 & 161.05 & 24.66 & 9.29 & 22.76 & \textbf{4.61} \\
InfiniHand & \underline{0.999} & \underline{0.998} & 23.42 & \underline{13.01} & \underline{5.10} & \underline{8.03} & \underline{4.96} & \underline{0.995} & \underline{0.982} & 45.83 & 20.79 & 7.90 & 9.48 & 8.78 \\
\midrule
POEM-v2 & 0.983 & 0.948 & \underline{22.35} & 14.52 & 7.72 & 9.11 & 20.88 & 0.983 & 0.950 & \underline{32.51} & \underline{16.65} & \underline{7.15} & 9.27 & 20.23 \\
Ours & \textbf{0.999} & \textbf{0.999} & \textbf{18.16} & \textbf{9.46} & \textbf{5.04} & \textbf{7.73} & \textbf{4.54} & \textbf{0.997} & \textbf{0.990} & \textbf{16.24} & \textbf{9.70} & \textbf{3.92} & \textbf{5.57} & \underline{5.98} \\
\bottomrule
\end{tabular}%
}
\end{table*}

%% file: figures/qualitative.tex
\begin{figure*}[t]
\centering
\includegraphics[width=\textwidth]{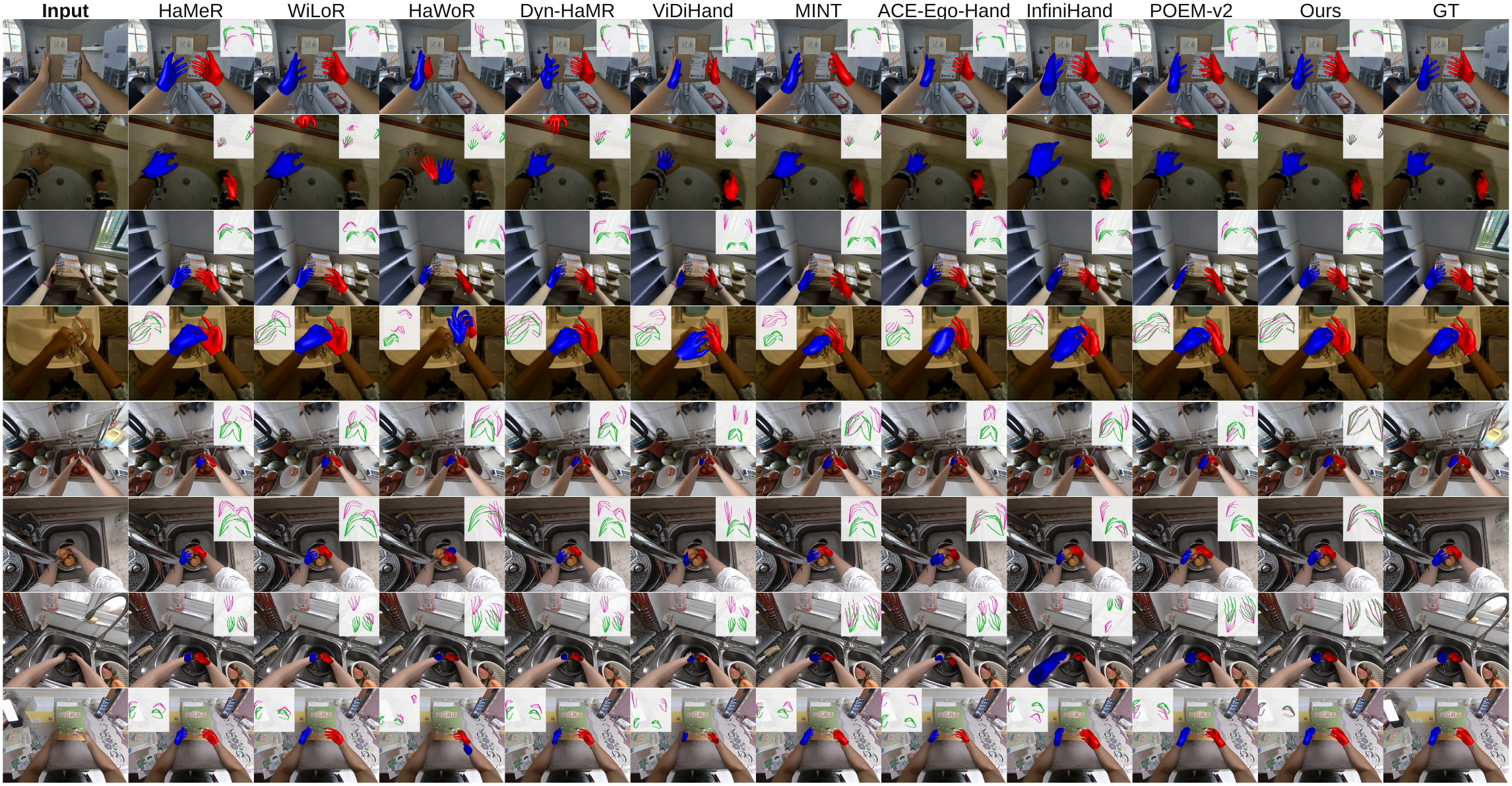}
\caption{\textbf{Qualitative comparison of camera-space hand reconstruction} on EgoSuite and GenHumanEgo. To jointly assess articulated pose and metric placement, image overlays show articulation, while camera X--Z insets make depth-placement differences relative to ground truth visible. Across these complementary views, StereoHand achieves the best 2D alignment and 3D placement.}
\label{fig:qualitative}
\end{figure*}

%% file: figures/world_qualitative.tex
\begin{figure*}[t]
\centering
\includegraphics[width=\textwidth]{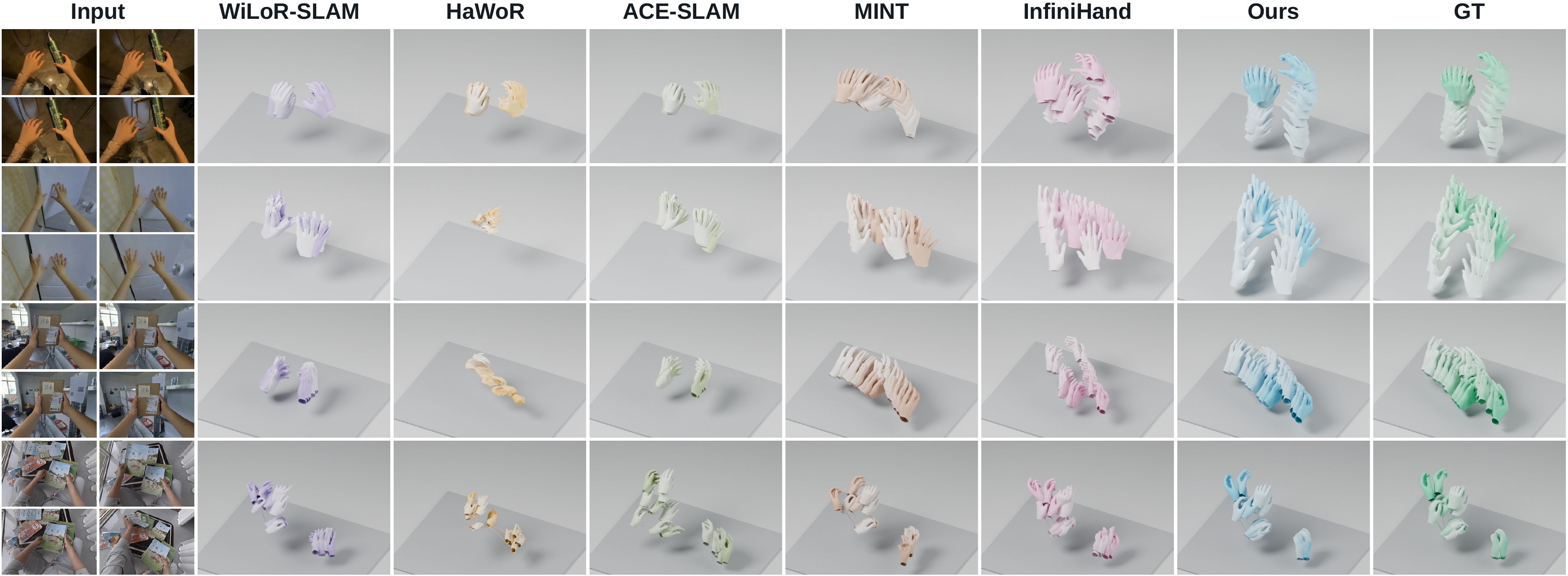}
\caption{\textbf{Qualitative comparison of world-space hand reconstruction on EgoSuite and GenHumanEgo.} Across both benchmarks, StereoHand reconstructs articulated hand poses and global trajectories that most closely match the ground truth.}
\label{fig:world_qualitative}
\end{figure*}

%% file: tables/world_hand.tex
\begin{table}[t]
\centering
\caption{\textbf{Quantitative comparison of world-space hand reconstruction} on EgoSuite and GenHumanEgo. StereoHand achieves the lowest errors across all metrics on both benchmarks.}
\label{tab:world_hand}
\scriptsize
\setlength{\tabcolsep}{4.5pt}
\renewcommand{\arraystretch}{1.1}
\resizebox{\columnwidth}{!}{%
\begin{tabular}{c@{\hspace{5pt}}lcccc}
\toprule
 & Method & W-MPJPE$\downarrow$ & WA-MPJPE$\downarrow$ & RTE$\downarrow$ & Accel$\downarrow$ \\
\midrule
\multirow{6}{*}{\rotatebox{90}{\fontsize{6}{7}\selectfont\textnormal{EgoSuite}}} & WiLoR-SLAM & 101.90 & 43.83 & 1.339 & 2.85 \\
 & HaWoR & 385.04 & 93.49 & 2.968 & 27.50 \\
 & ACE-SLAM & 94.65 & 44.24 & 0.632 & 2.29 \\
 & MINT & 101.95 & 46.13 & 0.763 & \underline{1.92} \\
 & InfiniHand & \underline{85.87} & \underline{41.02} & \underline{0.436} & 2.36 \\
 & Ours & \textbf{35.84} & \textbf{17.58} & \textbf{0.119} & \textbf{1.53} \\
\midrule
\multirow{6}{*}{\rotatebox{90}{\fontsize{6}{7}\selectfont\textnormal{GenHumanEgo}}} & WiLoR-SLAM & 48.67 & \underline{22.24} & 2.127 & 1.98 \\
 & HaWoR & 107.83 & 46.34 & 6.352 & 5.14 \\
 & ACE-SLAM & 49.76 & 24.14 & \underline{1.241} & 1.09 \\
 & MINT & 63.84 & 28.62 & 2.385 & 1.36 \\
 & InfiniHand & \underline{47.34} & 24.95 & 1.833 & \underline{1.09} \\
 & Ours & \textbf{26.55} & \textbf{13.39} & \textbf{0.688} & \textbf{0.90} \\
\bottomrule
\end{tabular}%
}
\end{table}

%% file: tables/slam_baselines.tex
\begin{table}[t]
\centering
\caption{\textbf{Quantitative comparison of camera tracking} on EgoSuite and GenHumanEgo. Our method achieves the most accurate pose estimation. $^{*}$ denotes incomplete trajectory coverage.}
\label{tab:slam_baselines}
\scriptsize
\setlength{\tabcolsep}{4pt}
\renewcommand{\arraystretch}{1.1}

\resizebox{\columnwidth}{!}{%
\begin{tabular}{lcccc}
\toprule
& \multicolumn{2}{c}{EgoSuite} & \multicolumn{2}{c}{GenHumanEgo}\\
\cmidrule(lr){2-3}\cmidrule(lr){4-5}
Method & ATE (cm)$\downarrow$ & RPE$_{\text{R}}$ ($^\circ$)$\downarrow$ & ATE (cm)$\downarrow$ & RPE$_{\text{R}}$ ($^\circ$)$\downarrow$\\
\midrule
MAC-VO & 21.13 & \underline{0.634} & 3.30 & \underline{0.147} \\
ORB-SLAM3 & 13.86$^{*}$ & 1.274$^{*}$ & 1.56$^{*}$ & 0.225$^{*}$ \\
VOCA & 22.77$^{*}$ & 1.832$^{*}$ & 5.04$^{*}$ & 0.406$^{*}$ \\
AirSLAM & 17.46$^{*}$ & 1.095$^{*}$ & 1.52$^{*}$ & 0.239$^{*}$ \\
DROID-SLAM & \underline{11.08} & 1.447 & \underline{0.56} & 0.215 \\
\midrule
Ours & \textbf{1.07} & \textbf{0.237} & \textbf{0.49} & \textbf{0.131} \\
\bottomrule
\end{tabular}%
}
\end{table}

%% file: tables/ablations.tex
\begin{table}[t]
\centering
\caption{\textbf{Quantitative comparison} on 30 EgoSuite and 30 GenHumanEgo scenes for camera-space ablation (top) and world-space ablation (bottom).}
\label{tab:ablations}
\input{tables/ablation_hand}
\par\vspace{5pt}
\input{tables/ablation_slam}
\par\vspace{2pt}
\end{table}

%% file: tables/ablation_hand.tex
\begingroup
\fontsize{6.5}{8}\selectfont
\setlength{\tabcolsep}{1pt}
\renewcommand{\arraystretch}{1.08}
\resizebox{\columnwidth}{!}{%
\begin{tabular}{@{}l@{\hspace{4pt}}cccccc@{}}
\toprule
Variant & FAcc$\uparrow$ & Abs-MPJPE$\downarrow$ & MPJPE$\downarrow$ & PA$\downarrow$ & $\mathrm{EPE}_{\text{2D}}\downarrow$ & Jitter$\downarrow$ \\
\midrule
\textbf{Full} & 0.994 & 17.16 & 9.51 & 4.37 & 6.08 & 4.99 \\
w/o right view & 0.937 & 40.19 & 16.40 & 5.97 & 9.62 & 5.95 \\
w/o stereo feat. & 0.995 & 30.31 & 11.78 & 5.26 & 8.04 & 5.52 \\
w/o reliability gating & 0.994 & 17.24 & 9.53 & 4.36 & 6.10 & 5.01 \\
w/ WiLoR bbox & 0.933 & 49.64 & 17.07 & 6.66 & 13.77 & 6.56 \\
\bottomrule
\end{tabular}%
}
\endgroup

%% file: tables/ablation_slam.tex
\begingroup
\fontsize{6.5}{8}\selectfont
\setlength{\tabcolsep}{1pt}
\renewcommand{\arraystretch}{1.08}
\resizebox{\columnwidth}{!}{%
\begin{tabular}{@{}l@{\hspace{4pt}}cccccc@{}}
\toprule
Variant & ATE$\downarrow$ & $\mathrm{RPE}_{\text{R}}\downarrow$ & W-MPJPE$\downarrow$ & WA-MPJPE$\downarrow$ & RTE$\downarrow$ & Accel$\downarrow$ \\
\midrule
\textbf{Full} & 0.86 & 0.220 & 25.62 & 15.08 & 0.236 & 1.148 \\
w/o descriptor & 1.42 & 0.243 & 30.79 & 16.15 & 0.257 & 1.189 \\
w/o global BA & 2.58 & 0.213 & 30.76 & 16.30 & 0.328 & 1.148 \\
w/o loop closure & 1.81 & 0.216 & 27.48 & 15.32 & 0.273 & 1.138 \\
w/o stereo constraints & 1.16 & 0.226 & 26.80 & 15.16 & 0.237 & 1.165 \\
\bottomrule
\end{tabular}%
}
\endgroup

%% file: sec/limitation.tex
\section{Limitations}
\label{sec:limitations}
Despite its gains in metric world-space hand reconstruction, \method{} remains subject to three main limitations. First, severe occlusion and motion blur obscure visual cues for localization and MANO estimation. While stereo provides complementary viewpoints, recovering fine articulation and metric positions remains challenging when both views lack sufficient evidence. Second, our framework assumes accurate stereo calibration and synchronization. Calibration errors distort cross-view geometry, whereas synchronization offsets induce inconsistent observations of moving content, both biasing hand localization and degrading camera tracking. Third, stereo SLAM relies on textured, static background regions for dense matching. Dominant foreground motion from hands or objects, weak textures, and motion blur can degrade static correspondences, causing tracking drift that propagates into world-space hand trajectories.

%% file: sec/conclusion.tex
\section{Conclusion}
\label{sec:conclusion}
We present \method, a unified framework for metric world-space hand motion reconstruction from egocentric stereo video. The framework brings together geometric and stereo foundation models to address two coupled challenges, locating the hands accurately in metric 3D space and recovering precise camera motion. Stereo-guided features support hand localization and MANO reconstruction, while dense matching drives stereo SLAM with local and global trajectory optimization. Evaluations on diverse egocentric sequences demonstrate improvements in hand reconstruction and camera tracking over monocular and stereo baselines, yielding more accurate world-space hand motion. These advances support more reliable 3D annotation of human activities for hand--object interaction reconstruction, human-to-robot motion retargeting, and manipulation policy learning. By preserving the spatial structure of human demonstrations, \method{} takes a step toward transforming abundant egocentric stereo video into geometric supervision for embodied intelligence.

%% file: sec/statement.tex
\section*{LLM Usage Statement}
\label{sec:statement}

In this work, we used generative AI tools solely to assist with translation, literature summarization and analysis, and manuscript editing to enhance readability. Conversely, we did not use generative AI tools for designing research methods and experiments, implementing methodologies, interpreting results, proposing or refining hypotheses, cleaning and reformatting datasets, or supporting qualitative and thematic data analysis. Generating synthetic datasets, proposing mathematical claims, providing key elements for proofs, and assisting in writing proofs are not applicable to this work. We thoroughly reviewed all AI-assisted outputs by manually cross-checking translated and polished text sentence by sentence to ensure the original intent remained uncompromised. Finally, we take full responsibility for the final content of this work, including all text, claims, and artifacts produced with the aid of generative AI.

%% file: sec/appendix.tex
This supplement covers evaluation metrics (Sec.~\ref{app:metric_details}), mask, MANO, and matching losses (Sec.~\ref{app:losses}), hand localization (Sec.~\ref{app:hand_localization}), and qualitative results for hand reconstruction and camera tracking (Sec.~\ref{app:qualitative}).

\section{Evaluation Metrics}
\label{app:metric_details}

\subsection{Camera-Space Hand Metrics}
\label{app:camera_hand_metrics}

\paragraph{Hand detection.}
We compute 2D bounding boxes from projected hand joints (enlarging ground-truth boxes by 10\%) and define a hand as visible when at least one joint projects inside the image with depth $> 0.01\,\mathrm{m}$. Per frame, predictions are matched to visible ground truth by maximizing IoU plus a $0.5$ handedness-consistency bonus. Matches with positive IoU and identical handedness yield true positives (TP), unmatched visible ground truths yield false negatives (FN), and remaining visible predictions yield false positives (FP). Accumulating counts across $T$ frames yields the F1 score and frame accuracy (FAcc):
\begin{equation}
\begin{aligned}
\mathrm{F1} &= \frac{2\mathrm{TP}}{2\mathrm{TP}+\mathrm{FP}+\mathrm{FN}},\\
\mathrm{FAcc} &= \frac{1}{T}\sum_{t=1}^{T}
  \mathbf{1}\big[\mathrm{FP}_t=0\ \land\ \mathrm{FN}_t=0\big].
\end{aligned}
\end{equation}

\paragraph{Joint errors.}
For each hand $i$ ($J=21$ joints, wrist $j=0$), let $\widehat{\mathbf J}^{c}_{i,j}$ and $\mathbf J^{c}_{i,j}$ be the estimated and ground-truth 3D joints (in meters) in the left-camera frame. The evaluated prediction $\widetilde{\mathbf J}^{c}_{i,j}$ equals $\widehat{\mathbf J}^{c}_{i,j}$ for detected hands, and falls back to the same-side mean MANO hand placed along the central camera ray at depth $0.5\,\mathrm{m}$ for missed visible hands. The absolute, wrist-relative, and Procrustes-aligned per-joint position errors (in millimeters) are defined as
\begin{equation}
\begin{aligned}
e_{\text{Abs}}(i) &= \frac{10^3}{J}\sum_j
  \big\|\widetilde{\mathbf J}^{c}_{i,j}-\mathbf J^{c}_{i,j}\big\|_2,\\
e_{\text{MP}}(i) &= \frac{10^3}{J}\sum_j
  \big\|\widetilde{\mathbf J}^{c}_{i,j,\text{rel}}-\mathbf J^{c}_{i,j,\text{rel}}\big\|_2,\\
e_{\text{PA}}(i) &= \frac{10^3}{J}\sum_j
  \big\|A_i(\widehat{\mathbf J}^{c}_{i,j})-\mathbf J^{c}_{i,j}\big\|_2.
\end{aligned}
\end{equation}
The wrist-relative coordinates are $\mathbf J^{c}_{i,j,\text{rel}}=\mathbf J^{c}_{i,j}-\mathbf J^{c}_{i,0}$, with the same definition for $\widetilde{\mathbf J}^{c}_{i,j,\text{rel}}$. The similarity transform $A_i$ is fitted to the matched hand instance. For missed detections, $e_{\text{PA}}(i)$ defaults to the root-mean-square (RMS) distance of the ground-truth joints from their centroid (in millimeters). Final metrics are averaged over all hand instances across scenes.

For the set of visible target joints $\mathcal U$, let $d_{i,j}$ denote the Euclidean pixel distance between the 2D projections of $\widetilde{\mathbf J}^{c}_{i,j}$ and $\mathbf J^{c}_{i,j}$ computed using the camera intrinsics of the corresponding frame. We cap each distance at the image diagonal $D_i$ and compute the mean 2D projection error as
\begin{equation}
\mathrm{EPE}_{\text{2D}}=\frac{1}{|\mathcal U|}
  \sum_{(i,j)\in\mathcal U}\min(d_{i,j},D_i).
\end{equation}
For consecutive true-positive triplets of the same hand, we compute temporal jitter from the second-order differences of the predicted 3D joints as
\begin{equation}
\mathrm{Jitter}=10^3\operatorname{mean}_{t,h,j}
\big\|\widehat{\mathbf J}^{c,h}_{t+1,j}
 -2\widehat{\mathbf J}^{c,h}_{t,j}+\widehat{\mathbf J}^{c,h}_{t-1,j}\big\|_2.
\end{equation}
where $h$ indexes the handedness (left or right). Jitter is reported in $\mathrm{mm}/\mathrm{frame}^2$ without normalization by elapsed time.

\subsection{World-Space Hand Metrics}
\label{app:world_hand_metrics}

Before transforming the predicted hands into world coordinates, we align each estimated camera trajectory to its ground-truth reference via a single sequence-level $\SE$ transformation without scale alignment. Following the HaWoR~\citep{hawor} protocol, jointly valid observations for each hand are partitioned into contiguous segments of at least 16 frames, where missing predictions break the continuity rather than incurring a penalty.

\paragraph{W-MPJPE and WA-MPJPE.}
Let $\mathcal W$ denote a non-overlapping block of 2--100 frames within a run, and $A_{\mathcal S}$ be the similarity transform fitted to the hand joints across frames $\mathcal S\subseteq\mathcal W$. The aligned joint error (in mm) is defined as
\begin{equation}
E(\mathcal W;\mathcal S)=\frac{10^3}{J|\mathcal W|}\sum_{t\in\mathcal W}\sum_j
  \big\|A_{\mathcal S}(\widehat{\mathbf J}^{w}_{t,j})-\mathbf J^{w}_{t,j}\big\|_2.
\end{equation}
where $w$ indicates world coordinates and the hand index is omitted for brevity. Specifically, W-MPJPE estimates $A_{\mathcal S}$ using only the first two frames, whereas WA-MPJPE fits it over the entire block. Finally, we aggregate the resulting per-frame errors across all hands and scenes.

\paragraph{Root trajectory error.}
For a given run $\mathcal R$, let $\widehat{\mathbf r}_t$ and $\mathbf r_t$ denote the estimated and target wrist positions at frame $t$, and let $G_{\mathcal R}$ be the fixed-scale rigid transformation fitted between their trajectories. The target path length and the normalized per-frame residual are defined as
\begin{equation}
\begin{aligned}
L_{\mathcal R} &= \sum_{t\in\mathcal R\setminus\{\min\mathcal R\}}
                \|\mathbf r_t-\mathbf r_{t-1}\|_2,\\
e_{\text{RTE}}(t) &= 100\,\frac{\|G_{\mathcal R}(\widehat{\mathbf r}_t)-\mathbf r_t\|_2}
                                {L_{\mathcal R}}.
\end{aligned}
\end{equation}
The RTE is computed by averaging these residuals across all valid frames as a percentage, excluding zero-length trajectories.

\paragraph{Acceleration error.}
For joint positions $\mathbf x_t$ sampled at timestamps $\tau_t$, let $\delta_t=\tau_{t+1}-\tau_t$ denote the time step and $\mathbf v_t=(\mathbf x_{t+1}-\mathbf x_t)/\delta_t$ be the finite-difference velocity. The nonuniform second-order finite-difference acceleration $\mathbf a_t$ is given by
\begin{equation}
\mathbf a_t(\mathbf x)=\frac{2(\mathbf v_t-\mathbf v_{t-1})}{\delta_t+\delta_{t-1}}.
\end{equation}
The acceleration error (Accel) averages $\|\mathbf a_t(\widehat{\mathbf J}^{w}_{:,j})-\mathbf a_t(\mathbf J^{w}_{:,j})\|_2$ across all joints and valid samples in $\mathrm{m}/\mathrm{s}^2$, discarding the first and last acceleration samples of each continuous run.

\subsection{Camera Pose Metrics}
\label{app:camera_pose_metrics}

Let $\widehat{\mathbf c}_t$ and $\mathbf c_t$ denote the estimated and ground-truth left-camera centers (in meters), respectively. Over the valid frame indices $\mathcal T$, let $G\in\SE$ be the least-squares rigid alignment mapping the estimated centers to the ground truth. ATE~\citep{sturm} measures the aligned translation RMSE in centimeters and is given by
\begin{equation}
\mathrm{ATE}=100\sqrt{\frac{1}{|\mathcal T|}\sum_{t\in\mathcal T}
  \big\|G(\widehat{\mathbf c}_t)-\mathbf c_t\big\|_2^2}.
\end{equation}
For each consecutive valid pair $(i,j)\in\mathcal P$, we compute the camera-to-world relative rotations $\Delta\mathbf R_{ij}=\mathbf R_i^\top\mathbf R_j$ and $\Delta\widehat{\mathbf R}_{ij}=\widehat{\mathbf R}_i^\top\widehat{\mathbf R}_j$. The relative rotation error is given by the geodesic angle $\theta_{ij}$ of $(\Delta\mathbf R_{ij})^\top\Delta\widehat{\mathbf R}_{ij}$, derived from its trace. We report the RMSE of these angles in degrees as
\begin{equation}
\mathrm{RPE}_{\text{R}}=\frac{180}{\pi}\sqrt{\frac{1}{|\mathcal P|}
  \sum_{(i,j)\in\mathcal P}\theta_{ij}^{2}}.
\end{equation}
Pairs separated by more than $1.5$ times the median sampling interval are excluded.

All metrics are averaged equally across the retained scenes. Our method, DROID-SLAM~\citep{droid}, and MAC-VO~\citep{macvo} cover all 150 scenes. After excluding cases with an ATE exceeding $100\,\mathrm{cm}$, ORB-SLAM3~\citep{orbslam3}, AirSLAM~\citep{airslam}, and VOCA~\citep{voca} retain $(61, 35)$, $(59, 38)$, and $(79, 48)$ scenes on EgoSuite and GenHumanEgo, respectively. The same scene subsets are used to evaluate $\mathrm{RPE}_{\text{R}}$. In the camera-tracking table, asterisks indicate methods with incomplete trajectory coverage.

\section{Loss Functions}
\label{app:losses}

Following the notation in the main paper, we use $s$ to index the stereo views and $h$ to index the hands. The mask, MANO, and matching heads are trained independently.

\subsection{Hand Mask Losses}

For a mask on pixel set $\Omega$, let $m_u\in\{0,1\}$ be the ground-truth label and $p_u$ the predicted foreground probability at pixel $u$. The predicted probability of the correct class is $p_u^\star=p_u$ when $m_u=1$ and $p_u^\star=1-p_u$ otherwise. We use binary cross-entropy for pixel classification and soft Dice for foreground overlap.
\begin{equation}
\begin{gathered}
\mathcal L_{\text{BCE}} = -\frac{1}{|\Omega|}\sum_{u\in\Omega}\log p_u^\star,\\
\mathcal L_{\text{Dice}} = 1-\frac{2\sum_u p_um_u+\epsilon}
  {\sum_u p_u+\sum_u m_u+\epsilon}.
\end{gathered}
\end{equation}
We set $\epsilon=1$ and average these two loss terms across all valid mask annotations. During temporal fine-tuning, we also enforce geometric mask consistency. Source pixels are backprojected using detached source geometry and reprojected into the target view via the calibrated transformation. The consistency loss is then computed exclusively over valid projections within the target image.

\subsection{Hand MANO Losses}

For brevity, we omit time and hand indices and denote predicted quantities with hats ($\widehat{\cdot}$). Let $\mathbf R_0$ be the global hand rotation in the left camera frame, and $\mathbf R_1,\ldots,\mathbf R_{15}$ be the local MANO joint rotations. We adopt the geodesic distance
\begin{equation}
d(\widehat{\mathbf R},\mathbf R)=\arccos\!\left(\operatorname{clip}_{\epsilon}\!\left(\frac{\operatorname{tr}(\widehat{\mathbf R}^{\top}\mathbf R)-1}{2}\right)\right),
\end{equation}
where $\operatorname{clip}_{\epsilon}(\cdot)$ clamps values to $[-1+\epsilon,1-\epsilon]$ with $\epsilon=10^{-7}$ for numerical stability. The parameter supervision losses are formulated as
\begin{equation}
\begin{aligned}
\mathcal L_{\text{orient}} &= d(\widehat{\mathbf R}_0,\mathbf R_0), \\
\mathcal L_{\text{pose}} &= \tfrac{1}{15}\sum_{j=1}^{15} d(\widehat{\mathbf R}_j,\mathbf R_j), \\
\mathcal L_{\text{shape}} &= \tfrac{1}{10}\|\widehat{\boldsymbol\beta}-\boldsymbol\beta\|_2^2, \quad
\mathcal L_{\text{trans}} = \tfrac{1}{3}\|\widehat{\mathbf w}^{L}-\mathbf w^{L}\|_2^2.
\end{aligned}
\end{equation}
Here, $\mathbf w^L$ and $\mathbf J_j^L$ denote the metric wrist and joint positions in the left camera frame, whereas the two 2D supervision targets, $\mathbf p_j^c$ and $\mathbf p_j^L$, are expressed in normalized crop and image coordinates, respectively. The joint and reprojection losses are
\begin{equation}
\begin{aligned}
\mathcal L_{\text{3D}} &= \frac{1}{3|\mathcal V_{\text{3D}}|}
  \sum_{j\in\mathcal V_{\text{3D}}}\|\widehat{\mathbf J}_j^L-\mathbf J_j^L\|_1, \\
\mathcal L_{\text{2D}} &= \frac{1}{|\mathcal V_{\text{2D}}|}
  \sum_{j\in\mathcal V_{\text{2D}}}
   \|\widehat{\mathbf p}_j^c-\mathbf p_j^c\|_1, \\
\mathcal L_{\text{reproj}} &= \frac{1}{|\mathcal V_{\text{reproj}}|}
  \sum_{j\in\mathcal V_{\text{reproj}}}
   \|\bar\pi_{\mathbf K^L}(\widehat{\mathbf J}_j^L)-\mathbf p_j^L\|_1.
\end{aligned}
\end{equation}
Specifically, the joint head directly regresses $\widehat{\mathbf p}_j^c$, while $\bar\pi_{\mathbf K^L}$ projects the decoded 3D joints onto the normalized image plane. Each $\mathcal V$ denotes the set of valid joints for its corresponding loss term.

\begin{figure*}[t]
  \centering
  \includegraphics[width=.98\textwidth]{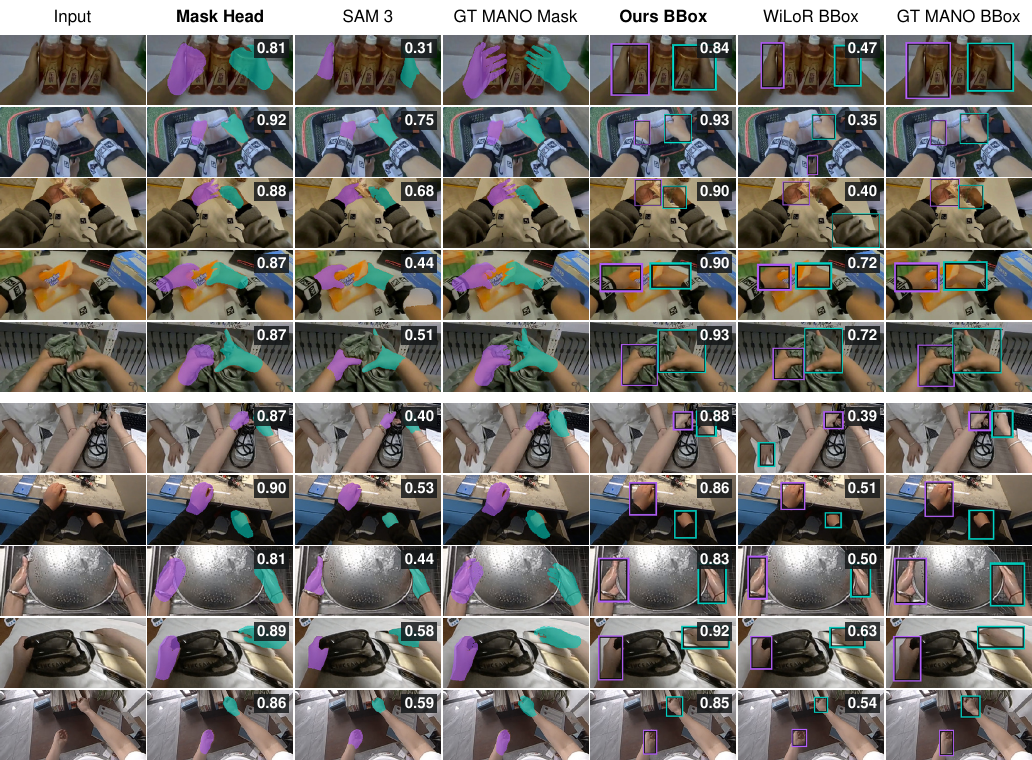}
  \caption{
  \textbf{Hand localization on EgoSuite~\citep{egosuite} (top five rows) and GenHumanEgo~\citep{genhumanego} (bottom five rows).}
On left-camera frames, we compare our masks with SAM~3~\citep{carion2026sam3} (\texttt{hand} prompt) and mask-derived boxes with WiLoR~\citep{wilor} detections. Purple and teal denote the left and right hands, respectively. Metrics show full-image IoU against projected MANO~\citep{mano} ground truth, evaluated on the two-hand silhouette union for masks and averaged per hand for boxes.}
  \label{fig:supp_hand_localization}
\end{figure*}

\subsection{Matching Head Losses}

We adopt the bidirectional InfoNCE objective from M$^3$~\citep{m3}. Given a reference image $I_1$ and a target image $I_k$, let $\widehat{\mathcal M}_k$ denote the set of ground-truth pixel correspondences observing the same 3D point, and let $\mathcal P_1$ and $\mathcal P_k$ be the respective candidate pixel sets. Using $\ell_2$-normalized feature descriptors $\mathbf D$ and an inverse temperature $\tau>0$, we define the pairwise similarity as $S_k(u,v)=\exp(\tau\mathbf D_{1,u}^{\top}\mathbf D_{k,v})$. The bidirectional loss for a positive pair $(u,v)\in\widehat{\mathcal M}_k$ is given by
\begin{equation}
\begin{aligned}
\ell_k(u,v)={}&-\log\frac{S_k(u,v)}{\sum_{w\in\mathcal P_1} S_k(w,v)} \\
 &-\log\frac{S_k(u,v)}{\sum_{w\in\mathcal P_k} S_k(u,w)}.
\end{aligned}
\end{equation}
Weighting each correspondence by its predicted confidence $q_{k,uv}>0$ and penalizing low confidence with a logarithmic term scaled by $\alpha>0$ gives
\begin{equation}
\begin{aligned}
\mathcal L_{\text{match}}={}&\frac{1}{N-1}\sum_{k=2}^{N}\sum_{(u,v)\in\widehat{\mathcal M}_k}
  \bigl[q_{k,uv}\,\ell_k(u,v) \\
  &\hspace{3em}-\alpha\log q_{k,uv}\bigr].
\end{aligned}
\end{equation}

\section{Hand Localization}
\label{app:hand_localization}

Our method derives the bounding box for MANO reconstruction from each hand's predicted silhouette. An off-center box shifts the viewing ray of the virtual crop camera, whereas an overly tight crop risks truncating the fingertips. To complement the quantitative detection metrics in Sec.~\ref{app:camera_hand_metrics}, Fig.~\ref{fig:supp_hand_localization} qualitatively compares our predicted masks against SAM~3~\citep{carion2026sam3} and our bounding boxes against WiLoR~\citep{wilor} on identical frames from the left camera.

The selected qualitative examples highlight two aspects of localization accuracy, spatial extent measured by box IoU and silhouette agreement measured by mask IoU. In the second EgoSuite example, our predicted bounding box achieves an IoU of $0.93$ compared to $0.35$ for WiLoR, whereas the disparity in mask IoU is less pronounced. The first GenHumanEgo example reveals a substantial gap in silhouette quality, where our method reaches a mask IoU of $0.87$ versus $0.40$ for SAM~3. A frame may still count as a correct detection even when its predicted box or mask is poorly aligned.

\section{Additional Qualitative Comparisons}
\label{app:qualitative}

\subsection{Camera-Space Hand Reconstruction}
We evaluate camera-space hand reconstruction on four in-domain EgoSuite scenes (Fig.~\ref{fig:supp_camera_hands}) and four out-of-domain GenHumanEgo scenes (Fig.~\ref{fig:supp_camera_hands_genhumanego}). Both figures pair 2D image overlays with unaligned $X$--$Z$ views to reveal depth errors that are otherwise masked by depth ambiguity in 2D projections. Overall, our method achieves superior reconstruction accuracy in both 2D alignment and metric 3D placement.

\subsection{World-Space Hand Motion}
Fig.~\ref{fig:supp_world_hands} illustrates qualitative comparisons of world-space hand motion across five in-domain EgoSuite sequences and five out-of-domain GenHumanEgo sequences. Compared to the baselines, our method more faithfully preserves local finger articulation as well as the relative spatial configuration between both hands. Furthermore, the reconstructed trajectories align more closely with the ground truth in both direction and spatial extent, demonstrating our framework's ability to accurately recover coherent bimanual motion in world coordinates.

\subsection{Camera Trajectories}
Reconstructing hand motion in world space depends jointly on accurate camera motion estimation and camera-space hand localization. Fig.~\ref{fig:slam_supplementary} compares the predicted camera trajectories against ground truth across six sequences. Compared to the baselines, our method tracks the ground-truth trajectories more closely across both turns and extended segments, demonstrating its ability to capture fine-grained local motion while preserving global trajectory consistency.

\input{figures/hand_supplementary}
\input{figures/slam_supplementary}

%% file: figures/hand_supplementary.tex
\begin{figure*}[p]
\centering
\includegraphics[width=.925\textwidth]{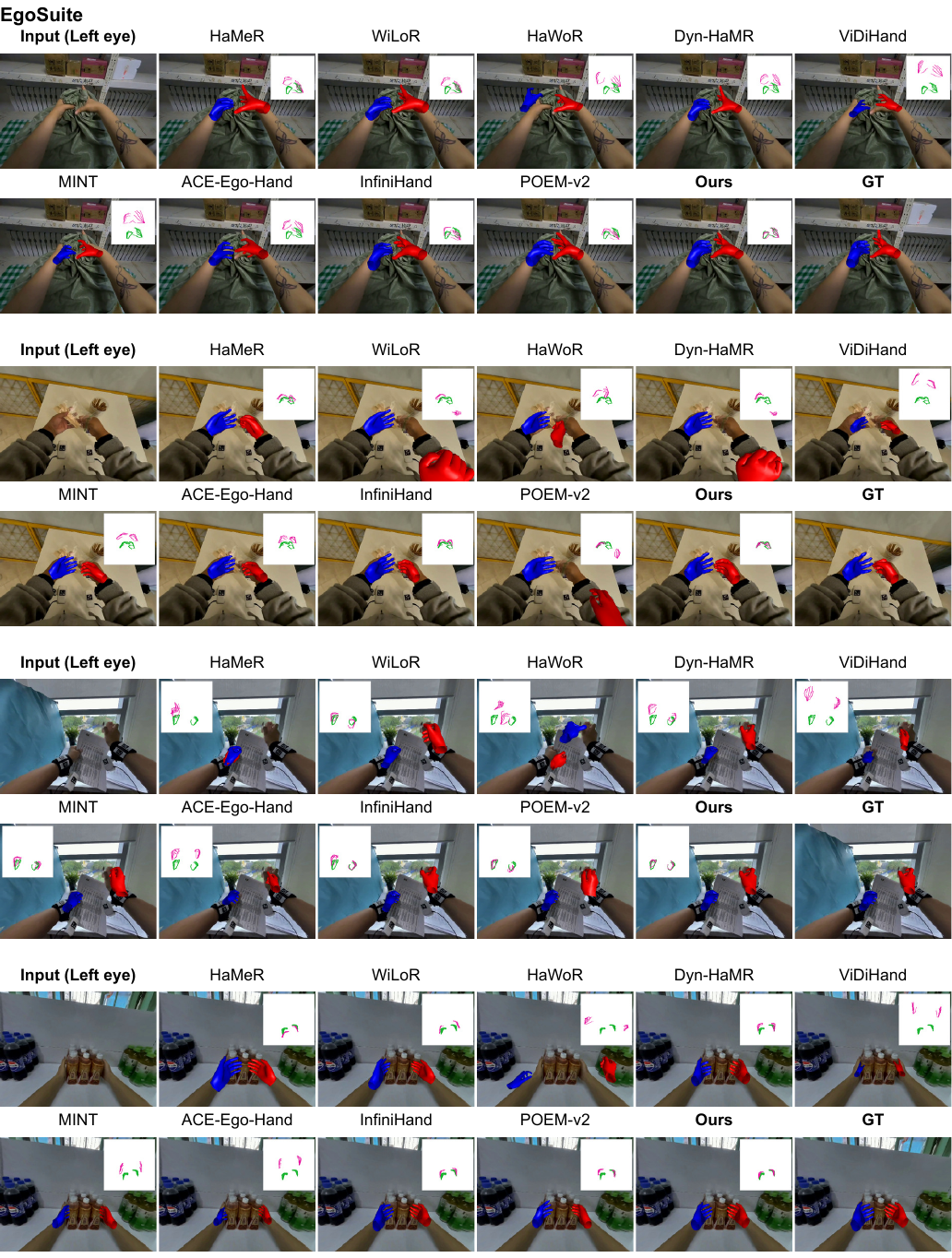}
\caption{\textbf{In-domain qualitative comparison on EgoSuite~\citep{egosuite}.} Image-space overlays illustrate hand articulation accuracy, while unaligned $X$--$Z$ views highlight depth offsets in camera space.}
\label{fig:supp_camera_hands}
\end{figure*}

\begin{figure*}[p]
\centering
\includegraphics[width=.925\textwidth]{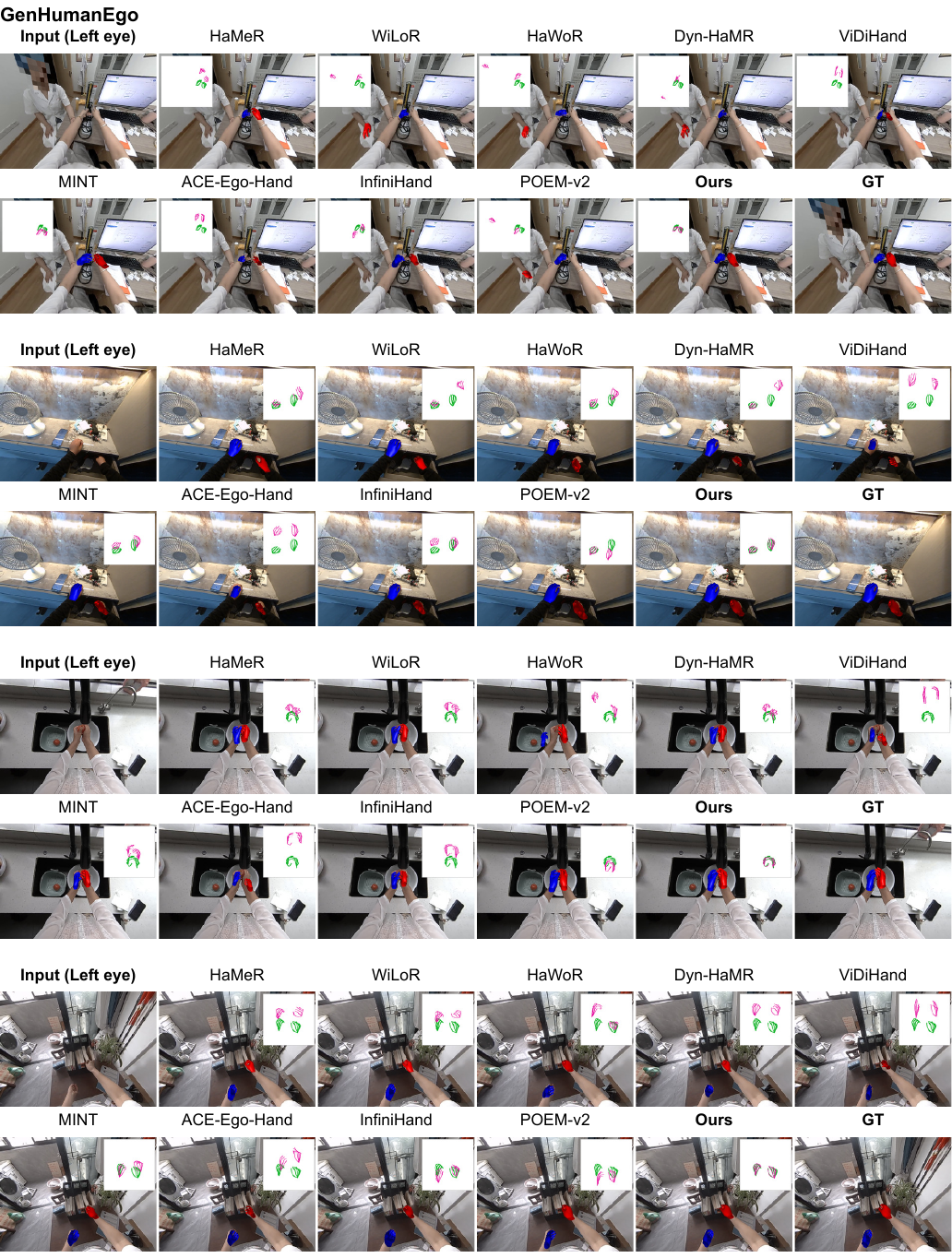}
\caption{\textbf{Out-of-domain qualitative comparison on GenHumanEgo~\citep{genhumanego}.} Image overlays illustrate hand articulation, while unaligned $X$--$Z$ views show metric hand placement in camera space.}
\label{fig:supp_camera_hands_genhumanego}
\end{figure*}

\begin{figure*}[p]
\centering
\includegraphics[width=\textwidth]{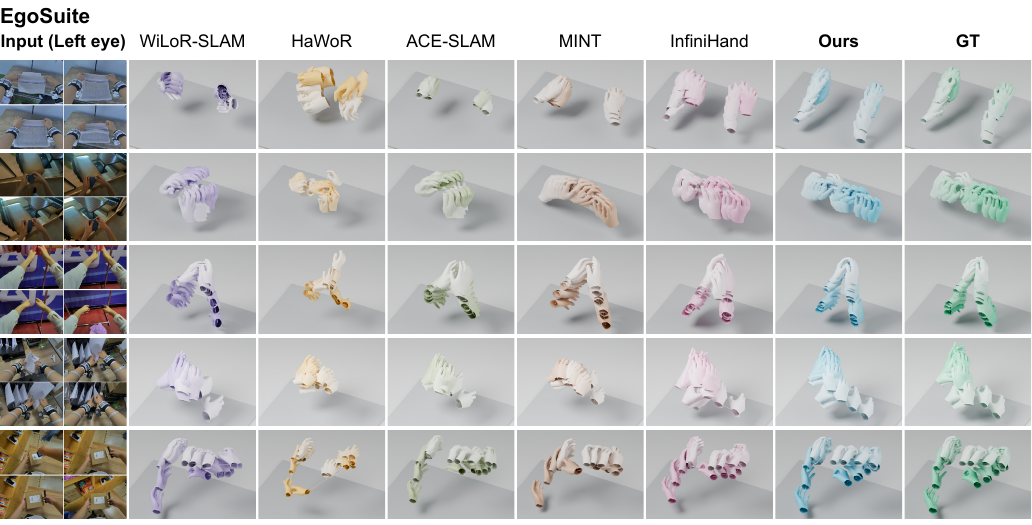}\par\vspace{8pt}
\includegraphics[width=\textwidth]{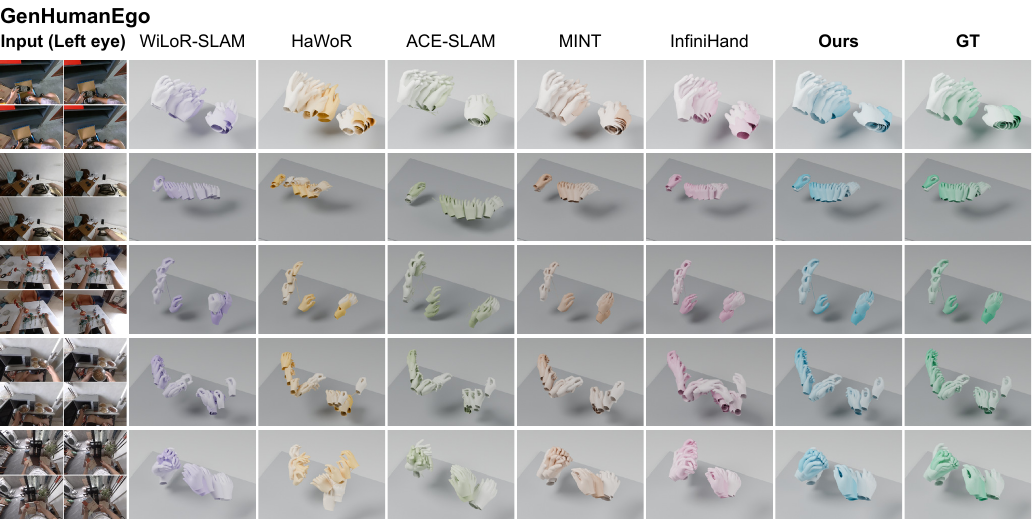}
\caption{\textbf{World-space hand motion on EgoSuite~\citep{egosuite} and GenHumanEgo~\citep{genhumanego}.}
Across five sequences from each benchmark, the leftmost column shows four input frames arranged in a $2\times2$ grid, while the remaining columns visualize the reconstructed world-space hand meshes at ten shared timestamps. For a fair comparison, each method is aligned via a single $\mathrm{Sim}(3)$ transformation fitted jointly across both hands.}
\label{fig:supp_world_hands}
\end{figure*}

%% file: figures/slam_supplementary.tex
\begin{figure*}[p]
\centering
\includegraphics[width=\textwidth,height=0.85\textheight,keepaspectratio]{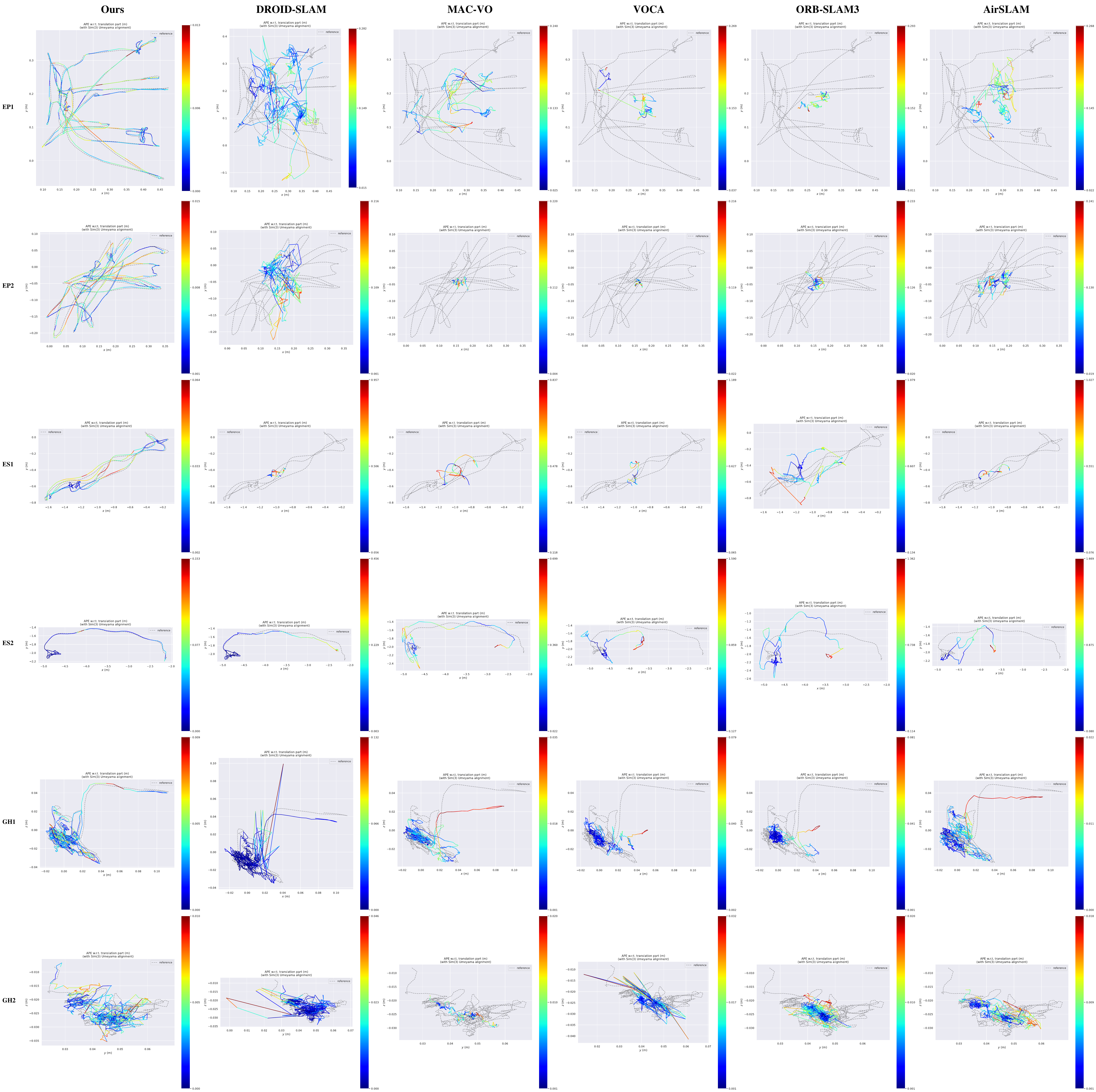}
\caption{\textbf{Camera trajectory comparisons on EgoSuite~\citep{egosuite} and GenHumanEgo~\citep{genhumanego}.}
Dashed curves represent the ground truth, while estimated trajectories are color-coded by translation APE after $\mathrm{Sim}(3)$ alignment.}
\label{fig:slam_supplementary}
\end{figure*}

%% file: references_stereohand.bib
@misc{pi3,
      title={$\pi^3$: Permutation-Equivariant Visual Geometry Learning}, 
      author={Yifan Wang and Jianjun Zhou and Haoyi Zhu and Wenzheng Chang and Yang Zhou and Zizun Li and Junyi Chen and Jiangmiao Pang and Chunhua Shen and Tong He},
      year={2026},
      eprint={2507.13347},
      archivePrefix={arXiv},
      primaryClass={cs.CV},
      url={https://arxiv.org/abs/2507.13347}, 
}

@misc{ffs,
      title={Fast-FoundationStereo: Real-Time Zero-Shot Stereo Matching}, 
      author={Bowen Wen and Shaurya Dewan and Stan Birchfield},
      year={2026},
      eprint={2512.11130},
      archivePrefix={arXiv},
      primaryClass={cs.CV},
      url={https://arxiv.org/abs/2512.11130}, 
}

@inproceedings{wilor,
  author    = {Rolandos Alexandros Potamias and Jinglei Zhang and Jiankang Deng and Stefanos Zafeiriou},
  title     = {{WiLoR}: End-to-end {3D} Hand Localization and Reconstruction in-the-wild},
  booktitle = {{IEEE/CVF} Conference on Computer Vision and Pattern Recognition ({CVPR})},
  pages     = {12242--12254},
  year      = {2025},
  doi       = {10.1109/CVPR52734.2025.01143},
}

@inproceedings{hamer,
      author       = {Georgios Pavlakos and
                      Dandan Shan and
                      Ilija Radosavovic and
                      Angjoo Kanazawa and
                      David Fouhey and
                      Jitendra Malik},
      title        = {Reconstructing Hands in 3D with Transformers},
      booktitle    = {{IEEE/CVF} Conference on Computer Vision and Pattern Recognition,
                      {CVPR} 2024, Seattle, WA, USA, June 16-22, 2024},
      pages        = {9826--9836},
      publisher    = {{IEEE}},
      year         = {2024},
      url          = {https://doi.org/10.1109/CVPR52733.2024.00938},
      doi          = {10.1109/CVPR52733.2024.00938},
      bibsource    = {dblp computer science bibliography, https://dblp.org}
    }

@misc{hawor,
      title={HaWoR: World-Space Hand Motion Reconstruction from Egocentric Videos}, 
      author={Jinglei Zhang and Jiankang Deng and Chao Ma and Rolandos Alexandros Potamias},
      year={2025},
      eprint={2501.02973},
      archivePrefix={arXiv},
      primaryClass={cs.CV},
      url={https://arxiv.org/abs/2501.02973}, 
}

@misc{dynhamr,
      title={Dyn-HaMR: Recovering 4D Interacting Hand Motion from a Dynamic Camera}, 
      author={Zhengdi Yu and Stefanos Zafeiriou and Tolga Birdal},
      year={2025},
      eprint={2412.12861},
      archivePrefix={arXiv},
      primaryClass={cs.CV},
      url={https://arxiv.org/abs/2412.12861}, 
}

@misc{poem,
      title={Multi-view Hand Reconstruction with a Point-Embedded Transformer}, 
      author={Lixin Yang and Licheng Zhong and Pengxiang Zhu and Xinyu Zhan and Junxiao Kong and Jian Xu and Cewu Lu},
      year={2026},
      eprint={2408.10581},
      archivePrefix={arXiv},
      primaryClass={cs.CV},
      url={https://arxiv.org/abs/2408.10581}, 
}

@misc{infinihand,
      title={InfiniHand: Streaming World-Space Hand Motion Estimation from Egocentric Video}, 
      author={Kerui Ren and Kaiwen Song and Weiguang Zhao and Yuxi Wang and Yufei Liu and Bo Dai and Haoyu Guo and Chunhua Shen and Mulin Yu and Tao Lu and Junting Dong},
      year={2026},
      eprint={2609.35743},
      archivePrefix={arXiv},
      primaryClass={cs.CV},
      url={https://arxiv.org/abs/2609.35743}, 
}

@inproceedings{droid,
      author       = {Zachary Teed and
                      Jia Deng},
      editor       = {Marc'Aurelio Ranzato and
                      Alina Beygelzimer and
                      Yann N. Dauphin and
                      Percy Liang and
                      Jennifer Wortman Vaughan},
      title        = {{DROID-SLAM:} Deep Visual {SLAM} for Monocular, Stereo, and {RGB-D}
                      Cameras},
      booktitle    = {Advances in Neural Information Processing Systems 34: Annual Conference
                      on Neural Information Processing Systems 2021, NeurIPS 2021, December
                      6-14, 2021, virtual},
      pages        = {16558--16569},
      year         = {2021},
      url          = {https://proceedings.neurips.cc/paper/2021/hash/89fcd07f20b6785b92134bd6c1d0fa42-Abstract.html},
      bibsource    = {dblp computer science bibliography, https://dblp.org}
    }

@misc{orbslam3,
      title={ORB-SLAM3: An Accurate Open-Source Library for Visual, Visual-Inertial and Multi-Map SLAM},
      author={Carlos Campos and Richard Elvira and Juan J. G\'{o}mez Rodr\'{i}guez and Jos\'{e} M. M. Montiel and Juan D. Tard\'{o}s},
      year={2021},
      eprint={2007.11898},
      archivePrefix={arXiv},
      primaryClass={cs.RO},
      url={https://arxiv.org/abs/2007.11898},
}

@misc{dpvo,
      title={Deep Patch Visual Odometry}, 
      author={Zachary Teed and Lahav Lipson and Jia Deng},
      year={2023},
      eprint={2208.04726},
      archivePrefix={arXiv},
      primaryClass={cs.CV},
      url={https://arxiv.org/abs/2208.04726}, 
}

@misc{m3,
      title={{M$^3$}: Dense Matching Meets Multi-View Foundation Models for Monocular Gaussian Splatting SLAM}, 
      author={Kerui Ren and Guanghao Li and Changjian Jiang and Yingxiang Xu and Tao Lu and Linning Xu and Junting Dong and Jiangmiao Pang and Mulin Yu and Bo Dai},
      year={2026},
      eprint={2603.16844},
      archivePrefix={arXiv},
      primaryClass={cs.CV},
      url={https://arxiv.org/abs/2603.16844}, 
}

@article{mano,
      author       = {Javier Romero and
                      Dimitrios Tzionas and
                      Michael J. Black},
      title        = {Embodied hands: modeling and capturing hands and bodies together},
      journal      = {{ACM} Trans. Graph.},
      volume       = {36},
      number       = {6},
      pages        = {245:1--245:17},
      year         = {2017},
      url          = {https://doi.org/10.1145/3130800.3130883},
      doi          = {10.1145/3130800.3130883},
      bibsource    = {dblp computer science bibliography, https://dblp.org}
    }

@misc{mast3rslam,
      title={MASt3R-SLAM: Real-Time Dense SLAM with 3D Reconstruction Priors}, 
      author={Riku Murai and Eric Dexheimer and Andrew J. Davison},
      year={2025},
      eprint={2412.12392},
      archivePrefix={arXiv},
      primaryClass={cs.CV},
      url={https://arxiv.org/abs/2412.12392}, 
}

@inproceedings{raftstereo,
      title={{RAFT-Stereo}: Multilevel Recurrent Field Transforms for Stereo Matching},
      author={Lahav Lipson and Zachary Teed and Jia Deng},
      booktitle={International Conference on 3D Vision (3DV)},
      year={2021},
      eprint={2109.07547},
      archivePrefix={arXiv},
      url={https://arxiv.org/abs/2109.07547},
}

@inproceedings{igevstereo,
      title={Iterative Geometry Encoding Volume for Stereo Matching},
      author={Gangwei Xu and Xianqi Wang and Xiaohuan Ding and Xin Yang},
      booktitle={Proceedings of the IEEE/CVF Conference on Computer Vision and Pattern Recognition},
      year={2023},
      eprint={2303.06615},
      archivePrefix={arXiv},
      url={https://arxiv.org/abs/2303.06615},
}

@inproceedings{foundationstereo,
      title={{FoundationStereo}: Zero-Shot Stereo Matching},
      author={Bowen Wen and Matthew Trepte and Joseph Aribido and Jan Kautz and Orazio Gallo and Stan Birchfield},
      booktitle={Proceedings of the IEEE/CVF Conference on Computer Vision and Pattern Recognition},
      year={2025},
      eprint={2501.09898},
      archivePrefix={arXiv},
      url={https://arxiv.org/abs/2501.09898},
}

@inproceedings{macvo,
      title={{MAC-VO}: Metrics-Aware Covariance for Learning-Based Stereo Visual Odometry},
      author={Yuheng Qiu and Yutian Chen and Zihao Zhang and Wenshan Wang and Sebastian Scherer},
      booktitle={IEEE International Conference on Robotics and Automation (ICRA)},
      pages={3803--3814},
      year={2025},
      url={https://arxiv.org/abs/2409.09479},
}

@article{airslam,
      title={{AirSLAM}: An Efficient and Illumination-Robust Point-Line Visual SLAM System},
      author={Kuan Xu and Yuefan Hao and Shenghai Yuan and Chen Wang and Lihua Xie},
      journal={IEEE Transactions on Robotics},
      volume={41},
      pages={1673--1692},
      year={2025},
      doi={10.1109/TRO.2025.3539171},
      url={https://arxiv.org/abs/2408.03520},
}

@inproceedings{voca,
      title={{VOCA}: Visual Odometry with Codec Awareness},
      author={Nouri Alexander Hilscher and Mateo de Mayo and Dominik Muhle and Christoph Otten genannt Hermes and Daniel Cremers},
      booktitle={European Conference on Computer Vision (ECCV)},
      year={2026},
      url={https://arxiv.org/abs/2607.00189},
}

@inproceedings{salad,
      title={Optimal Transport Aggregation for Visual Place Recognition},
      author={Sergio Izquierdo and Javier Civera},
      booktitle={Proceedings of the IEEE/CVF Conference on Computer Vision and Pattern Recognition},
      pages={17658--17668},
      year={2024},
      url={https://openaccess.thecvf.com/content/CVPR2024/html/Izquierdo_Optimal_Transport_Aggregation_for_Visual_Place_Recognition_CVPR_2024_paper.html},
}

@misc{vidihand,
      title={The Surprising Effectiveness of Video Diffusion Models for Hand Motion Reconstruction},
      author={Yuxi Wang and Chengkai Jin and Yufei Liu and Wenqi Ouyang and Tianyi Wei and Zhiwei Zeng and Siyuan Huang and Zhiqi Shen and Xingang Pan},
      year={2026},
      eprint={2606.30308},
      archivePrefix={arXiv},
      primaryClass={cs.CV},
      url={https://arxiv.org/abs/2606.30308},
}

@misc{mint,
      title={{MINT}: A Unified Model for World-Space Camera and Hand Motion Estimation from Scalable Egocentric Pipeline Supervision},
      author={Zijie Zhu and Weiren Cai and Yizhou Wang and Zhenjie Yang and Yide Liu and Jiahao Chen and Guanqi He},
      year={2026},
      eprint={2609.04958},
      archivePrefix={arXiv},
      primaryClass={cs.CV},
      url={https://arxiv.org/abs/2609.04958},
}

@misc{aceegohand,
      title={{ACE-Ego-Hand}: Repurposing Video Diffusion Models for Occlusion-Robust Egocentric 3D Hand Motion Recovery},
      author={Yufei Liu and Xixi Wang and Hao Li and Ganlong Zhao and Kaitong Cai and Chengkai Jin and Chunxiao Liu and Jianbo Liu and Siyuan Huang and Xingang Pan and Hongsheng Li},
      year={2026},
      eprint={2608.20308},
      archivePrefix={arXiv},
      primaryClass={cs.CV},
      url={https://arxiv.org/abs/2608.20308},
}

@inproceedings{hold,
  author    = {Zicong Fan and Maria Parelli and Maria Eleni Kadoglou and Xu Chen and Muhammed Kocabas and Michael J. Black and Otmar Hilliges},
  title     = {{HOLD}: Category-Agnostic {3D} Reconstruction of Interacting Hands and Objects from Video},
  booktitle = {Proceedings of the {IEEE/CVF} Conference on Computer Vision and Pattern Recognition},
  pages     = {494--504},
  year      = {2024},
  bibsource = {dblp computer science bibliography, https://dblp.org}
}

@inproceedings{dexmv,
  author    = {Yuzhe Qin and Yueh-Hua Wu and Shaowei Liu and Hanwen Jiang and Ruihan Yang and Yang Fu and Xiaolong Wang},
  title     = {{DexMV}: Imitation Learning for Dexterous Manipulation from Human Videos},
  booktitle = {Computer Vision -- {ECCV} 2022},
  pages     = {570--587},
  year      = {2022},
  doi       = {10.1007/978-3-031-19842-7_33},
  bibsource = {dblp computer science bibliography, https://dblp.org}
}

@article{egovla,
  author    = {Ruihan Yang and Qinxi Yu and Yecheng Wu and Rui Yan and Borui Li and An-Chieh Cheng and Xueyan Zou and Yunhao Fang and Xuxin Cheng and Ri-Zhao Qiu and Hongxu Yin and Sifei Liu and Song Han and Yao Lu and Xiaolong Wang},
  title     = {{EgoVLA}: Learning Vision-Language-Action Models from Egocentric Human Videos},
  journal   = {CoRR},
  volume    = {abs/2507.12440},
  year      = {2025},
  url       = {https://arxiv.org/abs/2507.12440},
  eprinttype = {arXiv},
  eprint    = {2507.12440},
  bibsource = {dblp computer science bibliography, https://dblp.org}
}

@article{skelwam,
  author    = {Zetao Cai and Yaping Li and Yiqun Wang and Xinyu Zhan and Yuyin Yang and Haoxiang Ma and Kailin Li and Tao Lu and Jiangmiao Pang and Linning Xu and Dahua Lin},
  title     = {{Skel-WAM}: A Hand-Skeleton-Conditioned World Action Model for Human-to-Robot Manipulation Transfer},
  journal   = {CoRR},
  volume    = {abs/2609.21514},
  year      = {2026},
  url       = {https://arxiv.org/abs/2609.21514},
  eprinttype = {arXiv},
  eprint    = {2609.21514},
  bibsource = {dblp computer science bibliography, https://dblp.org}
}

@inproceedings{liteanystereo,
      title={{Lite Any Stereo}: Efficient Zero-Shot Stereo Matching},
      author={Junpeng Jing and Weixun Luo and Ye Mao and Krystian Mikolajczyk},
      booktitle={Proceedings of the IEEE/CVF Conference on Computer Vision and Pattern Recognition},
      year={2026},
      eprint={2511.16555},
      archivePrefix={arXiv},
      primaryClass={cs.CV},
      url={https://arxiv.org/abs/2511.16555},
}

@misc{liteanystereov2,
      title={{Lite Any Stereo V2}: Faster and Stronger Efficient Zero-Shot Stereo Matching},
      author={Junpeng Jing and Ronglai Zuo and Zhelun Shen and Shangchen Zhou and Rolandos Alexandros Potamias and Stefanos Zafeiriou and Krystian Mikolajczyk and Jiankang Deng},
      year={2026},
      eprint={2606.24457},
      archivePrefix={arXiv},
      primaryClass={cs.CV},
      url={https://arxiv.org/abs/2606.24457},
}

@inproceedings{ego4d,
      author       = {Kristen Grauman and
                      Andrew Westbury and
                      Eugene Byrne and
                      Zachary Chavis and
                      Antonino Furnari and
                      Rohit Girdhar and
                      Jackson Hamburger and
                      Hao Jiang and
                      Miao Liu and
                      Xingyu Liu and
                      Miguel Martin and
                      Tushar Nagarajan and
                      Ilija Radosavovic and
                      Santhosh Kumar Ramakrishnan and
                      Fiona Ryan and
                      Jayant Sharma and
                      Michael Wray and
                      Mengmeng Xu and
                      Eric Zhongcong Xu and
                      Chen Zhao and
                      Siddhant Bansal and
                      Dhruv Batra and
                      Vincent Cartillier and
                      Sean Crane and
                      Tien Do and
                      Morrie Doulaty and
                      Akshay Erapalli and
                      Christoph Feichtenhofer and
                      Adriano Fragomeni and
                      Qichen Fu and
                      Abrham Gebreselasie and
                      Cristina Gonz{\'{a}}lez and
                      James Hillis and
                      Xuhua Huang and
                      Yifei Huang and
                      Wenqi Jia and
                      Weslie Khoo and
                      J{\'{a}}chym Kol{\'{a}}r and
                      Satwik Kottur and
                      Anurag Kumar and
                      Federico Landini and
                      Chao Li and
                      Yanghao Li and
                      Zhenqiang Li and
                      Karttikeya Mangalam and
                      Raghava Modhugu and
                      Jonathan Munro and
                      Tullie Murrell and
                      Takumi Nishiyasu and
                      Will Price and
                      Paola Ruiz Puentes and
                      Merey Ramazanova and
                      Leda Sari and
                      Kiran K. Somasundaram and
                      Audrey Southerland and
                      Yusuke Sugano and
                      Ruijie Tao and
                      Minh Vo and
                      Yuchen Wang and
                      Xindi Wu and
                      Takuma Yagi and
                      Ziwei Zhao and
                      Yunyi Zhu and
                      Pablo Arbel{\'{a}}ez and
                      David Crandall and
                      Dima Damen and
                      Giovanni Maria Farinella and
                      Christian Fuegen and
                      Bernard Ghanem and
                      Vamsi Krishna Ithapu and
                      C. V. Jawahar and
                      Hanbyul Joo and
                      Kris Kitani and
                      Haizhou Li and
                      Richard A. Newcombe and
                      Aude Oliva and
                      Hyun Soo Park and
                      James M. Rehg and
                      Yoichi Sato and
                      Jianbo Shi and
                      Mike Zheng Shou and
                      Antonio Torralba and
                      Lorenzo Torresani and
                      Mingfei Yan and
                      Jitendra Malik},
      title        = {Ego4D: Around the World in 3, 000 Hours of Egocentric Video},
      booktitle    = {{IEEE/CVF} Conference on Computer Vision and Pattern Recognition,
                      {CVPR} 2022, New Orleans, LA, USA, June 18-24, 2022},
      pages        = {18973--18990},
      publisher    = {{IEEE}},
      year         = {2022},
      url          = {https://doi.org/10.1109/CVPR52688.2022.01842},
      doi          = {10.1109/CVPR52688.2022.01842},
      bibsource    = {dblp computer science bibliography, https://dblp.org}
    }

@inproceedings{hot3d,
      author       = {Prithviraj Banerjee and
                      Sindi Shkodrani and
                      Pierre Moulon and
                      Shreyas Hampali and
                      Shangchen Han and
                      Fan Zhang and
                      Linguang Zhang and
                      Jade Fountain and
                      Edward Miller and
                      Selen Basol and
                      Richard A. Newcombe and
                      Robert Wang and
                      Jakob Julian Engel and
                      Tomas Hodan},
      title        = {{HOT3D:} Hand and Object Tracking in 3D from Egocentric Multi-View
                      Videos},
      booktitle    = {{IEEE/CVF} Conference on Computer Vision and Pattern Recognition,
                      {CVPR} 2025, Nashville, TN, USA, June 11-15, 2025},
      pages        = {7061--7071},
      publisher    = {Computer Vision Foundation / {IEEE}},
      year         = {2025},
      url          = {https://openaccess.thecvf.com/content/CVPR2025/html/Banerjee\_HOT3D\_Hand\_and\_Object\_Tracking\_in\_3D\_from\_Egocentric\_Multi-View\_CVPR\_2025\_paper.html},
      doi          = {10.1109/CVPR52734.2025.00662},
      bibsource    = {dblp computer science bibliography, https://dblp.org}
    }

@inproceedings{arctic,
      author       = {Zicong Fan and
                      Omid Taheri and
                      Dimitrios Tzionas and
                      Muhammed Kocabas and
                      Manuel Kaufmann and
                      Michael J. Black and
                      Otmar Hilliges},
      title        = {{ARCTIC:} {A} Dataset for Dexterous Bimanual Hand-Object Manipulation},
      booktitle    = {{IEEE/CVF} Conference on Computer Vision and Pattern Recognition,
                      {CVPR} 2023, Vancouver, BC, Canada, June 17-24, 2023},
      pages        = {12943--12954},
      publisher    = {{IEEE}},
      year         = {2023},
      url          = {https://doi.org/10.1109/CVPR52729.2023.01244},
      doi          = {10.1109/CVPR52729.2023.01244},
      bibsource    = {dblp computer science bibliography, https://dblp.org}
    }

@inproceedings{dexycb,
      author       = {Yu{-}Wei Chao and
                      Wei Yang and
                      Yu Xiang and
                      Pavlo Molchanov and
                      Ankur Handa and
                      Jonathan Tremblay and
                      Yashraj Narang and
                      Karl Van Wyk and
                      Umar Iqbal and
                      Stan Birchfield and
                      Jan Kautz and
                      Dieter Fox},
      title        = {DexYCB: {A} Benchmark for Capturing Hand Grasping of Objects},
      booktitle    = {{IEEE} Conference on Computer Vision and Pattern Recognition, {CVPR}
                      2021, virtual, June 19-25, 2021},
      pages        = {9044--9053},
      publisher    = {Computer Vision Foundation / {IEEE}},
      year         = {2021},
      url          = {https://openaccess.thecvf.com/content/CVPR2021/html/Chao\_DexYCB\_A\_Benchmark\_for\_Capturing\_Hand\_Grasping\_of\_Objects\_CVPR\_2021\_paper.html},
      doi          = {10.1109/CVPR46437.2021.00893},
      bibsource    = {dblp computer science bibliography, https://dblp.org}
    }

@inproceedings{meshgraphormer,
      author       = {Kevin Lin and
                      Lijuan Wang and
                      Zicheng Liu},
      title        = {Mesh Graphormer},
      booktitle    = {2021 {IEEE/CVF} International Conference on Computer Vision, {ICCV}
                      2021, Montreal, QC, Canada, October 10-17, 2021},
      pages        = {12919--12928},
      publisher    = {{IEEE}},
      year         = {2021},
      url          = {https://doi.org/10.1109/ICCV48922.2021.01270},
      doi          = {10.1109/ICCV48922.2021.01270},
      bibsource    = {dblp computer science bibliography, https://dblp.org}
    }

@inproceedings{interhand,
      author       = {Gyeongsik Moon and
                      Shoou{-}I Yu and
                      He Wen and
                      Takaaki Shiratori and
                      Kyoung Mu Lee},
      editor       = {Andrea Vedaldi and
                      Horst Bischof and
                      Thomas Brox and
                      Jan{-}Michael Frahm},
      title        = {InterHand2.6M: {A} Dataset and Baseline for 3D Interacting Hand Pose
                      Estimation from a Single {RGB} Image},
      booktitle    = {Computer Vision - {ECCV} 2020 - 16th European Conference, Glasgow,
                      UK, August 23-28, 2020, Proceedings, Part {XX}},
      series       = {Lecture Notes in Computer Science},
      volume       = {12365},
      pages        = {548--564},
      publisher    = {Springer},
      year         = {2020},
      url          = {https://doi.org/10.1007/978-3-030-58565-5\_33},
      doi          = {10.1007/978-3-030-58565-5\_33},
      bibsource    = {dblp computer science bibliography, https://dblp.org}
    }

@inproceedings{anyteleop,
      author       = {Yuzhe Qin and
                      Wei Yang and
                      Binghao Huang and
                      Karl Van Wyk and
                      Hao Su and
                      Xiaolong Wang and
                      Yu{-}Wei Chao and
                      Dieter Fox},
      editor       = {Kostas E. Bekris and
                      Kris Hauser and
                      Sylvia L. Herbert and
                      Jingjin Yu},
      title        = {AnyTeleop: {A} General Vision-Based Dexterous Robot Arm-Hand Teleoperation
                      System},
      booktitle    = {Robotics: Science and Systems XIX, Daegu, Republic of Korea, July
                      10-14, 2023},
      year         = {2023},
      url          = {https://doi.org/10.15607/RSS.2023.XIX.015},
      doi          = {10.15607/RSS.2023.XIX.015},
      bibsource    = {dblp computer science bibliography, https://dblp.org}
    }

@inproceedings{mast3r,
      author       = {Vincent Leroy and
                      Yohann Cabon and
                      J{\'{e}}r{\^{o}}me Revaud},
      editor       = {Ales Leonardis and
                      Elisa Ricci and
                      Stefan Roth and
                      Olga Russakovsky and
                      Torsten Sattler and
                      G{\"{u}}l Varol},
      title        = {Grounding Image Matching in 3D with MASt3R},
      booktitle    = {Computer Vision - {ECCV} 2024 - 18th European Conference, Milan, Italy,
                      September 29-October 4, 2024, Proceedings, Part {LXXII}},
      series       = {Lecture Notes in Computer Science},
      volume       = {15130},
      pages        = {71--91},
      publisher    = {Springer},
      year         = {2024},
      url          = {https://doi.org/10.1007/978-3-031-73220-1\_5},
      doi          = {10.1007/978-3-031-73220-1\_5},
      bibsource    = {dblp computer science bibliography, https://dblp.org}
    }

@inproceedings{dpt,
      author       = {Ren{\'{e}} Ranftl and
                      Alexey Bochkovskiy and
                      Vladlen Koltun},
      title        = {Vision Transformers for Dense Prediction},
      booktitle    = {2021 {IEEE/CVF} International Conference on Computer Vision, {ICCV}
                      2021, Montreal, QC, Canada, October 10-17, 2021},
      pages        = {12159--12168},
      publisher    = {{IEEE}},
      year         = {2021},
      url          = {https://doi.org/10.1109/ICCV48922.2021.01196},
      doi          = {10.1109/ICCV48922.2021.01196},
      bibsource    = {dblp computer science bibliography, https://dblp.org}
    }

@inproceedings{adamw,
      author       = {Ilya Loshchilov and
                      Frank Hutter},
      title        = {Decoupled Weight Decay Regularization},
      booktitle    = {7th International Conference on Learning Representations, {ICLR} 2019,
                      New Orleans, LA, USA, May 6-9, 2019},
      publisher    = {OpenReview.net},
      year         = {2019},
      url          = {https://openreview.net/forum?id=Bkg6RiCqY7},
      bibsource    = {dblp computer science bibliography, https://dblp.org}
    }

@inproceedings{ba,
      author       = {Bill Triggs and
                      Philip F. McLauchlan and
                      Richard I. Hartley and
                      Andrew W. Fitzgibbon},
      editor       = {Bill Triggs and
                      Andrew Zisserman and
                      Richard Szeliski},
      title        = {Bundle Adjustment - {A} Modern Synthesis},
      booktitle    = {Vision Algorithms: Theory and Practice, International Workshop on
                      Vision Algorithms, held during {ICCV} '99, Corfu, Greece, September
                      21-22, 1999, Proceedings},
      series       = {Lecture Notes in Computer Science},
      volume       = {1883},
      pages        = {298--372},
      publisher    = {Springer},
      year         = {1999},
      url          = {https://doi.org/10.1007/3-540-44480-7\_21},
      doi          = {10.1007/3-540-44480-7\_21},
      bibsource    = {dblp computer science bibliography, https://dblp.org}
    }

@book{hz,
      author       = {Richard Hartley and
                      Andrew Zisserman},
      title        = {Multiple View Geometry in Computer Vision},
      publisher    = {Cambridge University Press},
      year         = {2004},
      url          = {https://doi.org/10.1017/cbo9780511811685},
      doi          = {10.1017/CBO9780511811685},
      isbn         = {9780511811685},
      bibsource    = {dblp computer science bibliography, https://dblp.org}
    }

@article{orbslam2,
      author       = {Raul Mur{-}Artal and
                      Juan D. Tard{\'{o}}s},
      title        = {{ORB-SLAM2:} An Open-Source {SLAM} System for Monocular, Stereo, and
                      {RGB-D} Cameras},
      journal      = {{IEEE} Trans. Robotics},
      volume       = {33},
      number       = {5},
      pages        = {1255--1262},
      year         = {2017},
      url          = {https://doi.org/10.1109/TRO.2017.2705103},
      doi          = {10.1109/TRO.2017.2705103},
      bibsource    = {dblp computer science bibliography, https://dblp.org}
    }

@inproceedings{dust3r,
  author       = {Shuzhe Wang and
                  Vincent Leroy and
                  Yohann Cabon and
                  Boris Chidlovskii and
                  J{\'{e}}r{\^{o}}me Revaud},
  title        = {{DUSt3R:} Geometric 3D Vision Made Easy},
  booktitle    = {{IEEE/CVF} Conference on Computer Vision and Pattern Recognition,
                  {CVPR} 2024, Seattle, WA, USA, June 16-22, 2024},
  pages        = {20697--20709},
  publisher    = {{IEEE}},
  year         = {2024},
  url          = {https://doi.org/10.1109/CVPR52733.2024.01956},
  doi          = {10.1109/CVPR52733.2024.01956},
  bibsource    = {dblp computer science bibliography, https://dblp.org}
}

@article{vggt,
  author       = {Jianyuan Wang and
                  Minghao Chen and
                  Nikita Karaev and
                  Andrea Vedaldi and
                  Christian Rupprecht and
                  David Novotn{\'{y}}},
  title        = {{VGGT:} Visual Geometry Grounded Transformer},
  journal      = {CoRR},
  volume       = {abs/2503.11651},
  year         = {2025},
  url          = {https://doi.org/10.48550/arXiv.2503.11651},
  doi          = {10.48550/ARXIV.2503.11651},
  eprinttype   = {arXiv},
  eprint       = {2503.11651},
  bibsource    = {dblp computer science bibliography, https://dblp.org}
}

@inproceedings{metro,
  author       = {Kevin Lin and
                  Lijuan Wang and
                  Zicheng Liu},
  title        = {End-to-End Human Pose and Mesh Reconstruction with Transformers},
  booktitle    = {{IEEE} Conference on Computer Vision and Pattern Recognition, {CVPR}
                  2021, virtual, June 19-25, 2021},
  pages        = {1954--1963},
  publisher    = {Computer Vision Foundation / {IEEE}},
  year         = {2021},
  url          = {https://openaccess.thecvf.com/content/CVPR2021/html/Lin\_End-to-End\_Human\_Pose\_and\_Mesh\_Reconstruction\_with\_Transformers\_CVPR\_2021\_paper.html},
  doi          = {10.1109/CVPR46437.2021.00199},
  bibsource    = {dblp computer science bibliography, https://dblp.org}
}

@inproceedings{handoccnet,
  author       = {JoonKyu Park and
                  Yeonguk Oh and
                  Gyeongsik Moon and
                  Hongsuk Choi and
                  Kyoung Mu Lee},
  title        = {HandOccNet: Occlusion-Robust 3D Hand Mesh Estimation Network},
  booktitle    = {{IEEE/CVF} Conference on Computer Vision and Pattern Recognition,
                  {CVPR} 2022, New Orleans, LA, USA, June 18-24, 2022},
  pages        = {1486--1495},
  publisher    = {{IEEE}},
  year         = {2022},
  url          = {https://doi.org/10.1109/CVPR52688.2022.00155},
  doi          = {10.1109/CVPR52688.2022.00155},
  bibsource    = {dblp computer science bibliography, https://dblp.org}
}

@inproceedings{mobrecon,
  author       = {Xingyu Chen and
                  Yufeng Liu and
                  Yajiao Dong and
                  Xiong Zhang and
                  Chongyang Ma and
                  Yanmin Xiong and
                  Yuan Zhang and
                  Xiaoyan Guo},
  title        = {MobRecon: Mobile-Friendly Hand Mesh Reconstruction from Monocular
                  Image},
  booktitle    = {{IEEE/CVF} Conference on Computer Vision and Pattern Recognition,
                  {CVPR} 2022, New Orleans, LA, USA, June 18-24, 2022},
  pages        = {20512--20522},
  publisher    = {{IEEE}},
  year         = {2022},
  url          = {https://doi.org/10.1109/CVPR52688.2022.01989},
  doi          = {10.1109/CVPR52688.2022.01989},
  bibsource    = {dblp computer science bibliography, https://dblp.org}
}

@inproceedings{sturm,
  author = {J{\"u}rgen Sturm and Nikolas Engelhard and Felix Endres and Wolfram Burgard and Daniel Cremers},
  title = {A benchmark for the evaluation of {RGB-D SLAM} systems},
  booktitle = {2012 IEEE/RSJ International Conference on Intelligent Robots and Systems},
  pages = {573--580},
  year = {2012},
  url = {https://dblp.org/rec/conf/iros/SturmEEBC12},
  bibsource = {dblp computer science bibliography, https://dblp.org}
}

@misc{egosuite,
  author = {{Lightwheel}},
  title = {{EgoSuite-Open100K}: A Fully Annotated Open Egocentric Human Dataset},
  year = {2026},
  month = aug,
  howpublished = {\url{https://huggingface.co/collections/LightwheelAI/egosuite-open100k}},
  note = {Dataset}
}

@misc{genhumanego,
  title        = {{Gen-HumanEgo}},
  author       = {{GenRobot}},
  howpublished = {\url{https://huggingface.co/datasets/genrobot2025/Gen-HumanEgo}},
  year         = {2026},
}

@inproceedings{carion2026sam3,
  author    = {Nicolas Carion and Laura Gustafson and Yuan{-}Ting Hu and Shoubhik Debnath and Ronghang Hu and
               Didac Suris Coll{-}Vinent and Chaitanya Ryali and Kalyan Vasudev Alwala and Haitham Khedr and
               Andrew Huang and Jie Lei and Tengyu Ma and Baishan Guo and Arpit Kalla and Markus Marks and
               Joseph Greer and Meng Wang and Peize Sun and Roman R{\"{a}}dle and Triantafyllos Afouras and
               Effrosyni Mavroudi and Katherine Xu and Tsung{-}Han Wu and Yu Zhou and Liliane Momeni and
               Rishi Hazra and Shuangrui Ding and Sagar Vaze and Francois Porcher and Feng Li and Siyuan Li and
               Aishwarya Kamath and Ho Kei Cheng and Piotr Doll{\'{a}}r and Nikhila Ravi and Kate Saenko and
               Pengchuan Zhang and Christoph Feichtenhofer},
  title     = {{SAM 3}: Segment Anything with Concepts},
  booktitle = {The Fourteenth International Conference on Learning Representations ({ICLR})},
  year      = {2026},
  url       = {https://openreview.net/forum?id=r35clVtGzw},
}
